\pdfoutput=1

\documentclass[letterpaper, 10 pt, conference]{ieeeconf}
\IEEEoverridecommandlockouts
\usepackage{amsmath}
\usepackage{amssymb}
\usepackage{graphicx}
\usepackage{xcolor}
\usepackage{booktabs}
\usepackage{hyperref}
\hypersetup{hidelinks}
\usepackage{cite}
\usepackage{tikz}
\usetikzlibrary{arrows,plotmarks,calc,angles,quotes,decorations.markings,decorations.pathmorphing}

\tikzset{
    semithick/.style=  {line width=0.6pt}
}

\newcommand\centerofmass{
    \tikz[radius=0.37em] {
        \fill (0,0) -- ++(0.37em,0) arc [start angle=0,end angle=90] -- ++(0,-0.74em) arc [start angle=270, end angle=180];
        \fill [color=white] (0,0) -- ++(0,0.37em) arc [start angle=90,end angle=180] -- ++(0.74em,0) arc [start angle=0, end angle=-90];
        \draw (0,0) circle;
    }
}

\title{\LARGE \bf
Robust Control Synthesis and Verification for Wire-Borne Underactuated Brachiating Robots Using Sum-of-Squares Optimization
}

\author{Siavash Farzan$^{1}$, Ai-Ping Hu$^{2}$, Michael Bick$^{3}$, and Jonathan Rogers$^{4}$ %
\thanks{$^{1}$Institute for Robotics and Intelligent Machines, Georgia Institute of Technology, Atlanta, GA 30332, USA, e-mail: {\tt\footnotesize sfarzan@gatech.edu}}%
\thanks{$^{2}$Georgia Tech Research Institute, Atlanta, GA 30332, USA.}%
\thanks{$^{3}$Woodruff School of Mechanical Engineering, Georgia Institute of Technology, Atlanta, GA 30332, USA.}%
\thanks{$^{4}$Guggenheim School of Aerospace
Engineering, Georgia Institute of Technology, Atlanta, GA 30332, USA.}%
}

\begin{document}

\setlength{\abovedisplayskip}{3pt}
\setlength{\belowdisplayskip}{3pt}
\setlength{\abovedisplayshortskip}{3pt}
\setlength{\belowdisplayshortskip}{3pt}
\setlength{\textfloatsep}{3pt}
\setlength{\abovecaptionskip}{1pt}
\addtolength{\skip\footins}{-5pt}

\maketitle
\thispagestyle{empty}
\pagestyle{empty}

\begin{abstract}
Control of wire-borne underactuated brachiating robots requires a robust feedback control design that can deal with dynamic uncertainties, actuator constraints and unmeasurable states.
In this paper, we develop a robust feedback control for brachiating on flexible cables, building on previous work on optimal trajectory generation and time-varying LQR controller design. We propose a novel simplified model for approximation of the flexible cable dynamics, which enables inclusion of parametric model uncertainties in the system.
We then use semidefinite programming (SDP) and sum-of-squares (SOS) optimization to synthesize a time-varying feedback control with formal robustness guarantees to account for model uncertainties and unmeasurable states in the system. 
Through simulation, hardware experiments and comparison with a time-varying LQR controller, it is shown that the proposed robust controller results in relatively large robust backward reachable sets and is able to reliably track a pre-generated optimal trajectory and achieve the desired brachiating motion in the presence of parametric model uncertainties, actuator limits, and unobservable states.\looseness=-1
\end{abstract}

\section{INTRODUCTION AND RELATED WORK}
When considering mobile robots in practical settings, a key challenge is robust locomotion.
In unstructured environments ranging from cities to farmland, the ability of mobile robots to locomote safely in a robust manner independent of model constraints and uncertainties is at once both extremely important and extremely challenging.
The authors have developed a wire-borne underactuated brachiating robot \cite{Davies18} for potential applications such as 
precision agriculture, power line inspection, urban area surveillance, 
public safety, and traffic management.
However, traversing highly unstructured environments such as a network of elevated wires introduces a significant model uncertainty into the system,
as low-order deterministic equations of motion cannot capture the dynamics of flexible and deformable bodies. Other examples of wire-traversing robots have recently emerged, including the SlothBot \cite{NotEge19} capable of rolling on a mesh of wires. However, brachiating robots could offer unique advantages such as the capability to pass obstacles on wires, if they could overcome the challenge of swinging on such vibrating medium.\looseness=-1

Over the past two decades, research efforts on control of brachiating robots have exclusively focused on brachiation on rigid structures, such as ladders and monkey bars. The earliest brachiating robot was introduced by Fukuda \cite{FukKon91} for brachiation on ladder bars. Later, a heuristic control method was proposed for a two degrees of freedom robot locomoting on horizontal parallel bars \cite{SaiAra94}. The Target Dynamics algorithm was proposed in \cite{NakKod00} to enable locomotion of a simplified two-link brachiating robot over several rungs of a ladder. Using this method, instead of handling the system dynamics via reference trajectories, the control task is achieved by representing the robot dynamics with a simplified single pendulum as a lower dimensional target.
Zero-energy cost motions for passive brachiating models attached to an unchangeable ceiling were investigated in \cite{GomRui05}. An underactuated brachiating robot with magnetic ``feet'' was designed in \cite{MazAsa10}, for which a feedback linearization based controller was used to track optimal motion trajectories. 
In \cite{MegRah13}, a PD control and an adaptive robust control were employed to track optimal trajectories for a two link brachiating robot with uncertain kinematic and dynamic parameters, moving between fixed supports.
Pchelkin et al. \cite{PchGus16} presented an optimization framework to generate trajectories for energy efficient brachiation of a 24-DoF Gorilla robot on horizontal ladder bars.
A model-free sliding mode control scheme was presented in \cite{NguLiu17,NguLiu19} for brachiating
along a rigid structural member with an upward slope.
More recently, a three-link brachiation robot was presented in \cite{YanCho19}, which used an iterative LQR algorithm for trajectory generation and a combination of a cascaded PID control and an input-output linearization controller to track desired trajectories and swing along monkey bars.\looseness=-1

\begin{figure}[t]
	\centering
	\renewcommand{\arraystretch}{0.25}
    \setlength\tabcolsep{0.25pt}
    \begin{tabular}{cc}
	\includegraphics[trim={0bp 370bp 0bp 50bp},clip, width=0.625\columnwidth]{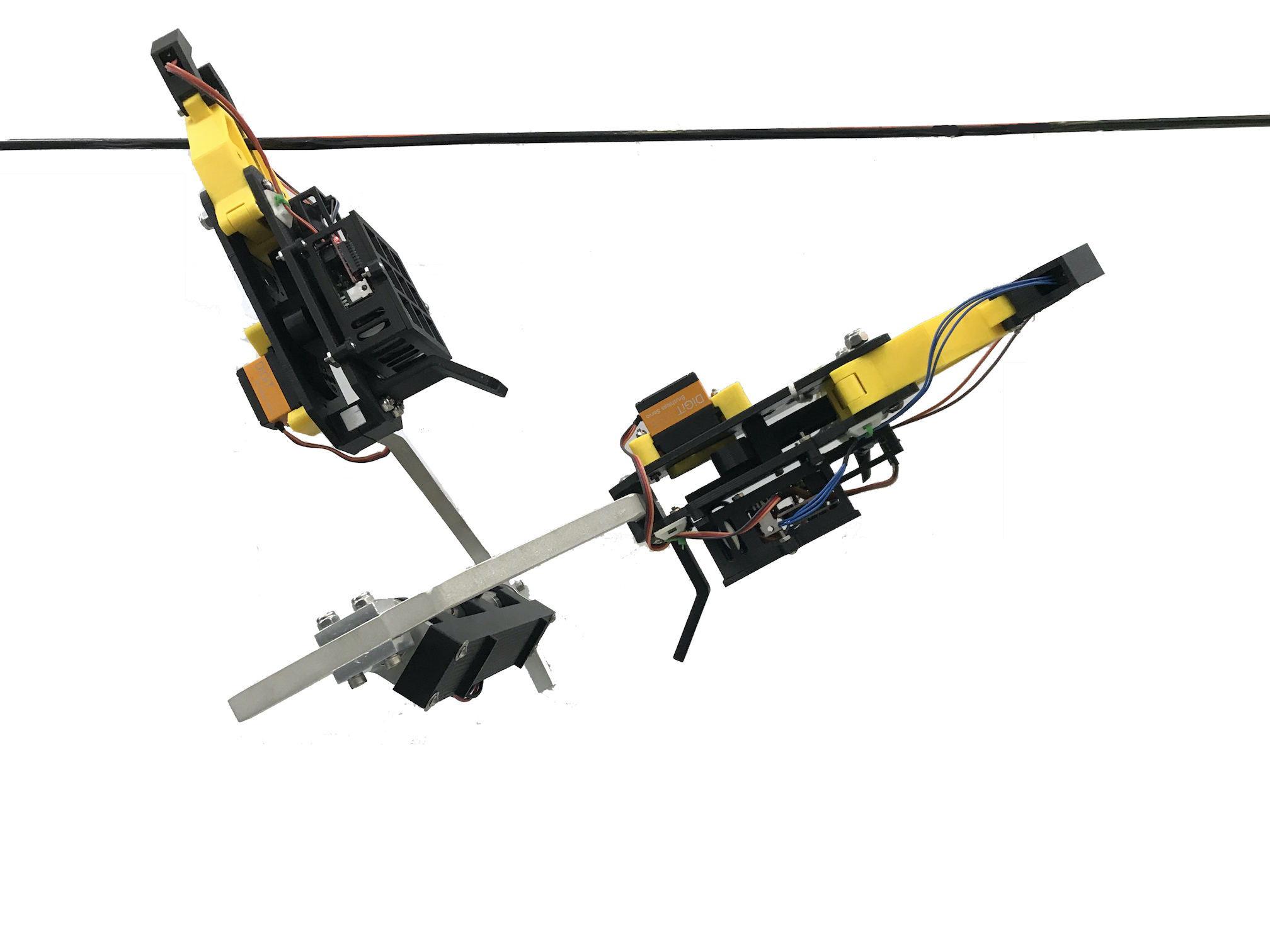} &
	\includegraphics[trim={0bp 0bp 0bp 0bp},clip, width=0.375\columnwidth]{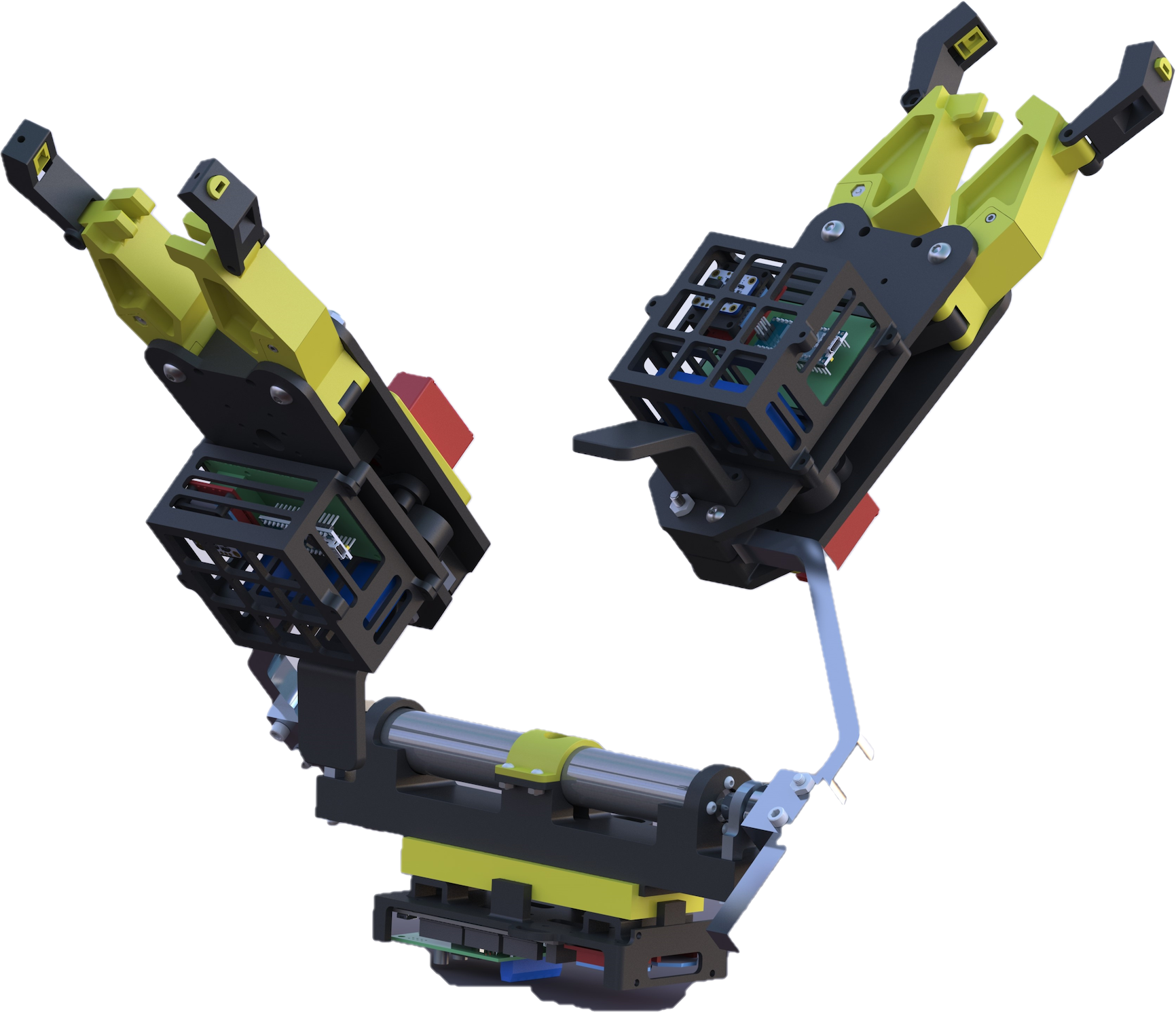} \\
	\scriptsize (a) & \scriptsize (b)
	\end{tabular}
	\caption{(a) Robot hardware prototype performing a brachiation maneuver, (b) CAD model of the robot prototype.}
	\label{fig:robot-hardware}
	\vspace{-2pt}
\end{figure}

To the best of the authors' knowledge, none of the prior works in the literature has addressed the problem of brachiating on a {vibrating} flexible cable. The authors have recently presented a time-varying LQR (TVLQR) controller design in \cite{Farzan19} for brachiation on flexible cables. However, the cable dynamics and its associated uncertainties could not be considered in the TVLQR design, and the region of attraction could not be formally verified for that controller. Building on previous work on optimal trajectory generation \cite{Farzan18} and TVLQR controller \cite{Farzan19}, this paper attempts to include the cable dynamics in the control design, and presents a robust closed-loop controller with formal guarantees for brachiating on flexible cables with uncertainties. We propose an approximate dynamic model
for the cable consisting of three parallel spring-dampers, enabling inclusion of parametric model uncertainties in the system.
Leveraging the recent developments in semidefinite programming (SDP) and sum-of-squares (SOS) optimization
\cite{Parrilo2003,BleTho12}, 
we synthesize a time-varying feedback control to account for
model uncertainties and unmeasurable states in the system,
and track pre-generated optimal trajectories to achieve a desired configuration.

In recent years, a large amount of work has been carried out on developing Lyapunov-based feedback controllers along with formal guarantees of their region of attraction (for time-invariant systems), or their invariant sets (for time-varying systems) via SOS
programming \cite{MajTed13,HenKor14,ManKui19,YinPac18,MajTed17},
which can accommodate external disturbances and model uncertainties in the dynamics.
These approaches can be mainly categorized into two methods: synthesizing closed-loop controllers while minimizing the outer approximation of the reachable sets \cite{MajTed17}, versus feedback control design by maximizing the inner approximation of the backward reachable sets \cite{MajTed13,YinPac18}.
While the former method is better suited for real-time planning in unknown environments, the latter provides the advantage of driving to a pre-defined goal from a larger set of initial conditions using a single reference trajectory.\looseness=-1
 
For the proposed wire-borne underactuated brachiating robot shown in Fig. \ref{fig:robot-hardware}, we compute an inner-approximation of the backward reachable set around a nominal trajectory for a given set of final configurations, and synthesize a feedback control action (in terms of the measurable states)
on a finite time horizon to maximize the size of the backward reachable set. The controller accounts for parametric model uncertainties, actuator limits, and unmeasurable states, while keeping motion trajectories inside the backward reachable set. The non-convex nonlinear optimization problem is formulated as a semidefinite program, which is broken down into approximate convex sub-problems and solved by an iterative algorithm using sum-of-squares programming.

In addition to the Lyapunov-based methods which employ sum-of-squares programming, there are other approaches in the controls literature for \emph{robust} control of underactuated systems, including adaptive control \cite{NguDan15,LiZha10,AziAme19}, sliding mode control \cite{NikFah07,AziSim18,Farzan-CDC20}
and backstepping \cite{HeGen10,KazMog13}. However, as will be shown throughout the paper, the SOS-based controller has the advantage of explicitly attempting to maximize the size of the robust backward reachable set, which results in a relatively large verified region for a single nominal trajectory. Moreover, the SOS-based controller is more robust to parametric uncertainties and control saturation, due to the fact that it directly accounts for bounded uncertainties and torque limits in the optimization process. Further, using SDP and SOS programming, there will be no need to design an observer for the unmeasurable states (the position and velocity of the gripper attached to the cable), as the control law can be constructed as an output feedback including only the measurable system states.

In summary, the main contributions of this work include: i) a novel dynamic modeling approach and a Fourier-based system identification to model the flexible cable as three parallel spring-dampers with parametric uncertainties, ii) a formally-verified robust feedback control design for brachiating robots, which takes the cable model uncertainties into consideration using sum-of-squares optimization, and iii) hardware validation of the proposed method by conducting real-world experiments on a brachiating robot prototype traversing a flexible cable. To our knowledge, the experimental result provides the first hardware evaluation of a feedback control design for brachiating robots attached to flexible cables. The proposed design also leads to the first SOS-based robust controller design in the domain of underactuated brachiating robots.\looseness=-1

\section{MULTI-BODY DYNAMIC MODEL}
\subsection{Flexible Cable Dynamics with Parametric Uncertainty}
Dynamics of a flexible cable with negligible bending and torsional stiffness can be described by partial differential equations (PDEs) \cite{AamFos00}. However, control of PDE models is challenging, as their solution is a function of both space and time and belongs to an infinite-dimensional space \cite{AhmPap16,GahPee17}.
One approach to analyze PDE systems relies on derivation of approximate models, for which the dynamic behavior of the system is described by ordinary differential equations (ODEs).
In \cite{Farzan18}, we presented
a robot-cable dynamic model consisting of a two-link underactuated brachiating robot, a lumped-mass flexible cable, and two soft junctions connecting the robot grippers to the cable.
The full-cable PDE system was approximated as a set of ODEs via a finite-element method \cite{BucFre04},
which resulted in a large number of generalized coordinates for the dynamic model, 
making it impractical to be included in a feedback control design.\looseness=-1

In this section, we propose a new approximate model which captures the dynamic effects of the flexible cable while keeping the model as a 3-DOF system described by ODEs. The proposed model provides the ability to include parametric model uncertainties in the state equations, paving the way for a ``robust'' feedback control design to compensate for discrepancies between the actual and the approximated models.\looseness=-1

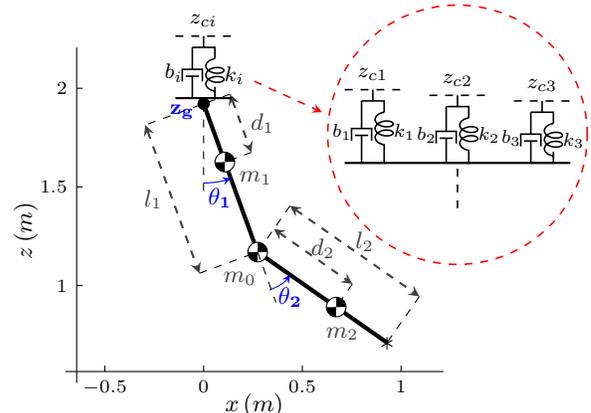
\begin{figure}[b]
\centering
\begin{tikzpicture}[scale=0.75,>=stealth]
    \tikzstyle{spring}=[semithick,decorate,decoration={aspect=0.55, segment length=1.2mm, amplitude=1.2mm,coil,mirror}]
    \tikzstyle{damper}=[semithick,decoration={markings,  
      mark connection node=dmp,
      mark=at position 0.5 with 
      {
        \node (dmp) [semithick,inner sep=0pt,transform shape,rotate=90,minimum width=7pt,minimum height=2pt,draw=none] {};
        \draw [semithick] ($(dmp.north east)+(-2pt,0)$) -- (dmp.south east) -- (dmp.south west) -- ($(dmp.north west)+(-2pt,0)$);
        \draw [semithick] ($(dmp.north)+(0,-2.5pt)$) -- ($(dmp.north)+(0,2.5pt)$);
      }
    }, decorate]
    \begin{scope}
    \pgfmathsetmacro{\AngleI}{-70}
    \pgfmathsetmacro{\AngleII}{-35}
    \coordinate (centro) at (2.25,4.75);
    \coordinate (centro2) at (6.75,3.5);
    \draw[ultra thick] (centro) -- ++(\AngleI:2.8) node(rod1) {} coordinate (joint1);
    \draw[ultra thick] (joint1) -- ++(\AngleII:2.8) node(rod2) {} coordinate (joint2);
    \filldraw [fill=black,draw=black] (centro) circle[radius=0.1];
    \pgfsetplotmarksize{0.7ex}
    \pgfplothandlermark{\pgfuseplotmark{asterisk}}
    \pgfplotstreamstart
    \pgfplotstreampoint{\pgfpoint{5.5cm}{0.5cm}}
    \pgfplotstreamend
    \path (centro) +(\AngleI:1.1) coordinate (com1);
    \path (joint1) +(\AngleII:1.7) coordinate (com2);
    \node [] at (joint1) {\centerofmass};
    \node [] at (com1) {\centerofmass};
    \node [] at (com2) {\centerofmass};
    \node at (joint1) [xshift=3pt,yshift=-4pt,darkgray] [below left] {\small $m_{0}$};
    \node at (com1) [xshift=2pt] [below right,darkgray] {\small $m_{1}$};
    \node at (com2) [xshift=12pt,yshift=-4pt,darkgray] [below left] {\small $m_{2}$};
    \draw (-0.2,0) -- (6.5,0);
    \draw (0,-0.2) -- (0,5.75);
    \draw[dashed,-] (centro) -- ++ (0,-1.6) node (mary1) [below]{};
    \draw[dashed,-] (joint1) -- +(\AngleI:0.95) node (mary2) [below]{};
    \path (centro) +(90+\AngleI:0.5) coordinate (d11);
    \path (com1) +(90+\AngleI:0.5) coordinate (d12);
    \path (joint1) +(90+\AngleII:0.5) coordinate (d21);
    \path (com2) +(90+\AngleII:0.5) coordinate (d22);
    \path (centro) +(90+\AngleI:-1.1) coordinate (l11);
    \path (joint1) +(90+\AngleI:-1.1) coordinate (l12);
    \path (joint1) +(90+\AngleII:1) coordinate (l21);
    \path (joint2) +(90+\AngleII:1) coordinate (l22);
    \draw[<->,dashed,thick,darkgray] (d11) -- (d12) node [midway,right]{\small $d_1$};
    \draw[<->,dashed,thick,darkgray] (d21) -- (d22) [xshift=5pt,yshift=4pt] node [midway]{\small $d_2$};
    \draw[<->,dashed,thick,darkgray] (l11) -- (l12) node [midway,left]{\small $l_1$};
    \draw[<->,dashed,thick,darkgray] (l21) -- (l22) [xshift=5pt,yshift=5pt] node [midway]{\small $l_2$};
    \draw[dashed] (d11) -- (l11);
    \draw[dashed] (d12) -- (com1);
    \draw[dashed] (l12) -- (joint1);
    \draw[dashed] (l21) -- (joint1);
    \draw[dashed] (d22) -- (com2);
    \draw[dashed] (l22) -- (joint2);
    \pic [draw, ->, "\small \color{blue} $\mathbf{\theta_1}$", angle eccentricity=1.2, angle radius=30, blue] {angle = mary1--centro--rod1};
    \pic [draw, ->, "\small \color{blue} $\mathbf{\theta_2}$", angle eccentricity=1.4, angle radius=15, blue] {angle = mary2--joint1--rod2};
    \node at (centro) [left,xshift=0pt,yshift=-3pt] {\footnotesize \color{blue} $\mathbf{z_{g}}$};
    \foreach \x/\xtext in {0.5/-0.5, 2.25/0, 4/0.5, 5.75/1}
      \draw[shift={(\x,0)}] (0pt,3pt) -- (0pt,0pt) node[below] {\scriptsize $\xtext$};
    \foreach \y/\ytext in {1.525/1 , 3.275/1.5, 5.025/2}
      \draw[shift={(0,\y)}] (3pt,0pt) -- (0pt,0pt) node[left] {\scriptsize $\ytext$};
   \end{scope}
   \begin{scope}
    \filldraw [fill=white,draw=white] (centro)++(0.9,0.4) circle[radius=0.01] coordinate (g);
    \filldraw [fill=white,draw=red,dashed,semithick] (g)++(3.6,-0.95) circle[radius=2.3];
    \draw [thin,->,dashed,red,semithick] (g) -- ++(1.2,-0.55);
   \end{scope}
   \begin{scope}
    \draw [semithick] (centro)++(0,1.2) -- ++(0,-0.2) -- ++(0.2,0) -- ++(0,-0.2) coordinate (p1);
    \draw [spring] (p1) -- ++(0,-0.6) node [midway,right]{\footnotesize $k_i$};
    \draw [semithick] (centro)++(0,0.1) -- ++(0.2,0) -- ++(0,0.2);
    \draw [semithick] (centro)++(0,0.1) -- ++(-0.2,0) coordinate (d1) ++(0,0.9) coordinate (d2);
    \draw [damper] (d1) -- ($(d1)!(d2)!(d1)$) node [midway,left]{\footnotesize $b_i$};
    \draw [semithick] (d2)  -- ++(0.2,0);
    \draw [semithick,dashed] (centro)++(-0.5,1.2)  -- ++(1,0) node [midway,above]{\footnotesize $z_{ci}$};
    \draw [thick] (centro)++(-0.5,0.1)  -- ++(1,0);
   \end{scope}
   \begin{scope}
    \draw [semithick] (centro2)++(-1.5,1.5) -- ++(0,-0.2) -- ++(0.2,0) -- ++(0,-0.2) coordinate (p1);
    \draw [spring] (p1) -- ++(0,-0.7) node [midway,right]{\scriptsize $k_1$};
    \draw [semithick] (centro2)++(-1.5,0.2) -- ++(0.2,0) -- ++(0,0.2);
    \draw [semithick] (centro2)++(-1.5,0.2) -- ++(-0.2,0) coordinate (d1) ++(0,1.1) coordinate (d2);
    \draw [damper] (d1) -- ($(d1)!(d2)!(d1)$) node [midway,left]{\scriptsize $b_1$};
    \draw [semithick] (d2)  -- ++(0.2,0);
    \draw [semithick,dashed] (centro2)++(-2,1.5)  -- ++(1,0) node [midway,above]{\footnotesize $z_{c1}$};
    \draw [semithick] (centro2)++(0,1.4) -- ++(0,-0.2) -- ++(0.2,0) -- ++(0,-0.2) coordinate (p1);
    \draw [spring] (p1) -- ++(0,-0.6) node [midway,right]{\scriptsize $k_2$};
    \draw [semithick] (centro2)++(0,0.2) -- ++(0.2,0) -- ++(0,0.2);
    \draw [semithick] (centro2)++(0,0.2) -- ++(-0.2,0) coordinate (d1) ++(0,1.0) coordinate (d2);
    \draw [damper] (d1) -- ($(d1)!(d2)!(d1)$) node [midway,left]{\scriptsize $b_2$};
    \draw [semithick] (d2)  -- ++(0.2,0);
    \draw [semithick,dashed] (centro2)++(-0.5,1.4)  -- ++(1,0) node [midway,above]{\footnotesize $z_{c2}$};
    \draw [semithick] (centro2)++(1.5,1.3) -- ++(0,-0.2) -- ++(0.2,0) -- ++(0,-0.2) coordinate (p1);
    \draw [spring] (p1) -- ++(0,-0.6) node [midway,right]{\scriptsize $k_3$};
    \draw [semithick] (centro2)++(1.5,0.2) -- ++(0.2,0) -- ++(0,0.2);
    \draw [semithick] (centro2)++(1.5,0.2) -- ++(-0.2,0) coordinate (d1) ++(0,0.9) coordinate (d2);
    \draw [damper] (d1) -- ($(d1)!(d2)!(d1)$) node [midway,left]{\scriptsize $b_3$};
    \draw [semithick] (d2)  -- ++(0.2,0);
    \draw [semithick,dashed] (centro2)++(1,1.3)  -- ++(1,0) node [midway,above]{\footnotesize $z_{c3}$};
    \draw [thick] (centro2)++(-2.0,0.2)  -- ++(4.0,0);
    \draw [semithick,dashed] (centro2)++(0,+0.1) -- ++(0,-0.75);
    \node at (3.15,-0.6) {\small $x\,(m)$};
    \node at (-0.9,2.5) [rotate=90] {\small $z\,(m)$};
   \end{scope}
\end{tikzpicture}
    \vspace{-0.1in}
    \caption{Multi-body model of the two-link brachiating robot with the proposed cable model consisting of three parallel spring-dampers.}
    \label{fig:model}
\end{figure}

The Fourier analysis of the vibration of the full-cable model during a brachiating maneuver reveals that only the first three harmonics contribute significantly to the dynamics (see Section \ref{subsec:sys-id}).
Thus, the vibrating effects of the flexible cable on the robot can be
adequately
captured by three parallel linear springs and dampers connecting the pivot gripper to three different attachment heights. The stiffness and damping coefficients of the spring-dampers, as well as the heights of the attachment points, depend on the physical characteristics of the cable, specifically the frequencies and the amplitudes of the harmonics retrieved by Fourier analysis.

The proposed model for the system, from which the system dynamics are obtained, is shown in Fig. \ref{fig:model}. The system consists of two rotational and one translational degrees of freedom (DOF):
$\theta_1$ as the angle of the first link with respect to the vertical axis, $\theta_2$ as the joint angle relative to the first link, and $z_g$ representing the vertical Cartesian position of the robot's gripper that is attached to the cable. The system has underactuation degree of 2, in the sense that only $\theta_2$ is actuated, with the robot's single torque actuator located at the joint between the two links.

We denote the stiffness, damping and attachment height of the three spring-dampers by $k_i$, $b_i$ and $z_{c_i}$ (for $i=1,2,3$) respectively. The parametric uncertainties in the cable model are taken into account by including an uncertainty term $w$ in the stiffness parameters of the three springs, that is
\begin{equation}
    k_i = k_{0_i} \pm w\,k_{0_i}, \quad i=1,2,3 \label{eq:stiffness}
\end{equation}
where $k_{0_i}$ denotes the nominal stiffness of the springs.

The nonlinear equations of motion of the system are derived by the Lagrangian method, with the state vector represented by
$x=[\theta_1,\,\theta_2,\,z_g,\dot{\theta}_1,\,\dot{\theta}_2,\,\dot{z}_g]^T$.

\vspace{-2pt}
\subsection{System Identification via Fourier Analysis}\label{subsec:sys-id}
\begin{figure}[t]
\renewcommand{\arraystretch}{0.25}
\setlength\tabcolsep{1pt}
\begin{tabular}{cc}
\includegraphics[trim={5bp 2bp 51bp 42bp},clip,width=0.525\columnwidth]{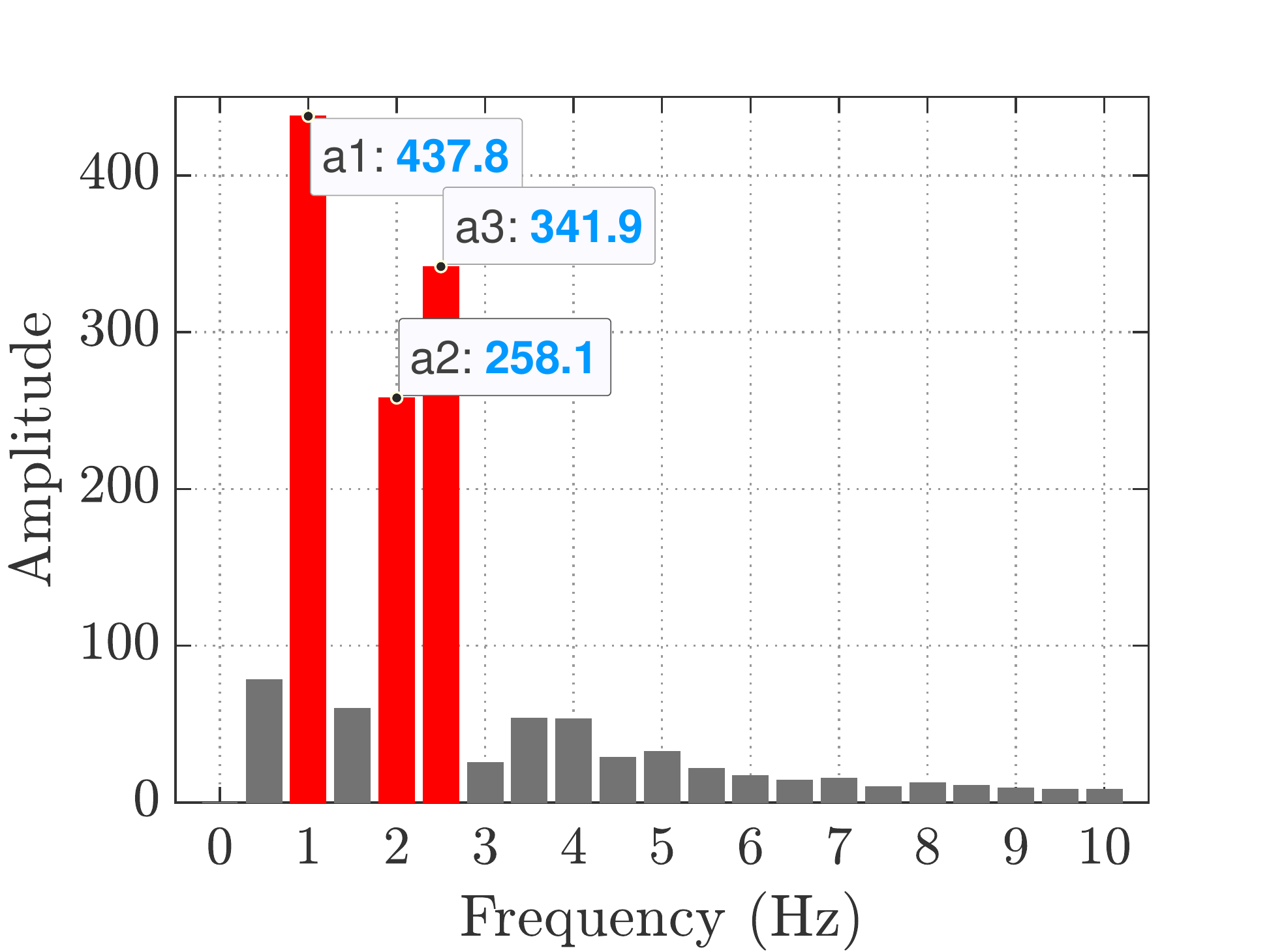} &
\includegraphics[trim={70bp 2bp 51bp 42bp},clip,width=0.457\columnwidth]{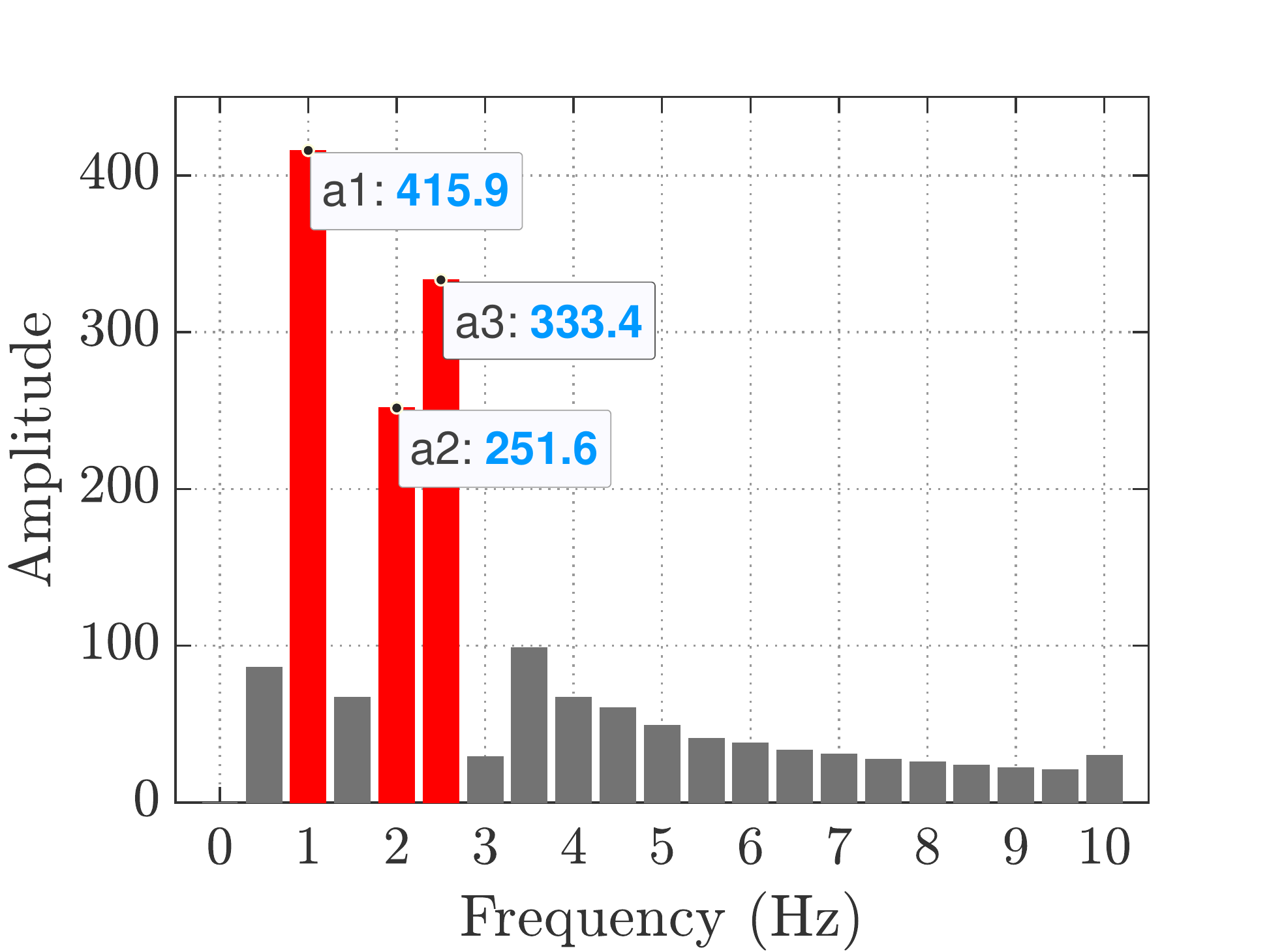}
\tabularnewline
{\small{}(a)} & {\small{}(b)}
\tabularnewline
\end{tabular}
\caption{\label{fig:fourier} Frequency spectrum of the cable vibration, (a) Finite-element model, (b) Proposed model. The first three harmonics are shown in red.}
\vspace{-2pt}
\end{figure}

\vspace{-2pt}
We use the output-error method \cite{jategaonkar2015flight}
to replicate the dynamic behavior of the full-cable with the proposed spring-dampers model. With the output-error method, the unknown system parameters including the stiffness, damping and attachment height of the three spring-dampers are tuned so that the Fourier spectrum of the position of the pivot gripper during a brachiating maneuver with the proposed model fits the Fourier spectrum of the robot with the full-cable model.
The algorithm for the output-error method can be summarized as: i) apply the same control input $u(t)$ to the robot-cable system with the full-cable model and the spring-dampers model, ii) compare the Fourier spectrum of the resulting simulated state $z_g$ using each model, iii) optimize the set of parameters $\{k_i,\,b_i,\,z_{c_i}\}$ (as the optimization decision variables) until the resulting Fourier spectrums -- in terms of the frequencies ($f_i,\,\hat{f}_i$) and the amplitudes ($a_i,\,\hat{a}_i$) of the first three harmonics -- are as close as possible in least squares sense:\looseness=-1
\begin{equation}
    \underset{k_i,\,b_i,\,z_{c_i}}{\textrm{argmin}}\;\;J=\sum_{i=1}^{3}(\hat{f}_i-{f}_i)^2+(\hat{a}_i-{a}_i)^2\textrm{,} \label{eq:output-error}
\end{equation}
where $J$ is the least square cost. 
We used the built-in MATLAB gradient-based optimization routine ``fmincon'' to solve the constrained nonlinear optimization problem in (\ref{eq:output-error}).
The system identification results are listed in Table \ref{tab:sys-id}, and the resulting Fourier spectrums
of the two models
are compared in Fig. \ref{fig:fourier}. The physical parameters of the full-cable model used as the reference will be shown later in Table \ref{tab:physical-parameters}.\looseness=-1

\begin{table}[t]
    \vspace{-2pt}
    \caption{System ID results for the proposed cable model.}
    \label{tab:sys-id}
    \centering
    \begin{tabular}{cccc}
    \toprule
    Spring \# & Stiffness $(k_0)$ & Damping $(b)$ & Attachment Height $(z_c)$ \\
    \midrule
    1 & 76.74 & 4.25 & 2.00 \\
    2 & 180.50 & 4.72 & 2.04 \\
    3 & 279.14 & 4.88 & 2.06 \\
    \bottomrule
    \end{tabular}
\end{table}

\section{CONTROL SYNTHESIS AND VERIFICATION}
With the parametric uncertainties present in the model described above, a robust closed-loop control design is required to control the robot brachiating on flexible cables. Moreover, the position and velocity of the pivot gripper attached to the cable ($z_g,\,\dot{z}_g$) cannot be measured in practice to be included in a state-feedback control policy.
We use semidefinite optimization and sum-of-squares programming to synthesize a time-varying feedback control law in terms of the measurable states of the system, with formal robustness guarantees against parametric uncertainties in dynamics and actuator saturations, in order to track a pre-generated optimal trajectory to a desired configuration.

\vspace{-1pt}
\subsection{Problem Formulation}
For the two-link brachiating robot attached to a flexible cable with a parametric model uncertainty, the nonlinear time-varying closed-loop equations of motion have the form:
\begin{equation}
    \hspace{-2pt} \dot{\bar{x}}(t)=f_{cl}(\bar{x},\bar{u},w,t), \;\, \bar{x}(t) \in \mathbb{R}^6, \; \bar{u}(\bar{y},t)\in\mathbb{R}, \; w\in\mathbb{R}\textrm{,} \label{eq:eom}
\end{equation}
with the state vector $\bar{x}(t)$ as the joint angles/velocities and the gripper position/velocity, and $\bar{y}\in\mathbb{R}^4$ as the output vector of the system formed by the measurable states:
\begin{gather}
    \bar{x}(t)=[\bar{\theta}_1(t),\,\bar{\theta}_2(t),\,\bar{z}_g(t),\,\dot{\bar{\theta}}_1(t),\,\dot{\bar{\theta}}_2(t),\,\dot{\bar{z}}_g(t)]^T\textrm{,} \\
    \bar{y}(t)=[\bar{\theta}_1(t),\,\bar{\theta}_2(t),\,\dot{\bar{\theta}}_1(t),\,\dot{\bar{\theta}}_2(t)]^T\textrm{.}
\end{gather}
The control input $\bar{u}(\bar{y},t)$ is the torque input at the center joint described by a time-varying feedback control in terms of the measurable states constrained by actuator saturation: $u_{min} \leq \bar{u}(\bar{y},t) \leq  u_{max}$.
Note that to derive the time-varying system equations, the states and the control input are defined as the deviations from a nominal reference trajectory:
\begin{equation}
    \bar{x}(t)=x(t)-x_{ref}(t), \quad \bar{u}(\bar{y},t)=u(\bar{y},t)-u_{ref}(t)\textrm{.}
\end{equation}

The uncertainty term $w$ is the parametric model uncertainty in the cable as in (\ref{eq:stiffness}), which is bounded and is described by the set $\mathcal{W}=\{w\in\mathbb{R}\;|\; w_{lb} \leq w \leq  w_{ub}\}$.

Defining the inner-approximation of the backward reachable set $\mathcal{B}_r(t)$ as a time-varying level set of the Lyapunov function $V(\bar{x},t)$:
\begin{equation}
    \mathcal{B}_r(t)=\{\bar{x} \in \mathbb{R}^6 \; | \; V(\bar{x},t)\leq r(t)\}\textrm{,}
\end{equation}
our objective is that given a set of desired final conditions $\mathcal{X}_f$, synthesize a time-varying output feedback controller $\bar{u}(\bar{y},t)$ that maximizes the volume of the set $\mathcal{B}_r(t)$ for any valid parametric model uncertainty, so that:
\begin{equation}
    \hspace{-3pt}\bar{x}(t_0) \in \mathcal{B}_r(t) \; \Rightarrow \; \bar{x}(t_f)\in \mathcal{X}_f, \; \forall t\in[t_0,t_f], \; \forall w\in\mathcal{W}\textrm{.} \label{eq:invariance}
\end{equation}
Equation (\ref{eq:invariance}) implies that if the robot initial configuration lies in the set $\mathcal{B}_r(t)$, it will be driven to the desired set $\mathcal{X}_f$ by the controller.\looseness=-1

To guarantee the invariance condition in (\ref{eq:invariance}), it is sufficient to insure that on the boundary of the set $\mathcal{B}_r(t)$, the Lyapunov function $V(\bar{x},t)$ increases by a slower rate than the boundary level, keeping the trajectories inside the level set, that is

\noindent \small\begin{equation}
    \hspace{-0.08in} V(\bar{x},t)=r(t) \Rightarrow
    \dot{V}(\bar{x},\bar{u},w,t)<\dot{r}(t), \forall t\in[t_0,t_f], \forall w\in \mathcal{W} \label{eq:const}
\end{equation} \normalsize

\noindent where the time derivative $\dot{V}$ is calculated by:
\begin{equation}
    \dot{V}(\bar{x},\bar{u},w,t)=\frac{\partial V(\bar{x},t)}{\partial \bar{x}}f_{cl}(\bar{x},\bar{u},w,t)+\frac{\partial V(\bar{x},t)}{\partial t}\textrm{.} \label{eq:vdot}
\end{equation}

By approximating the volume of the set $\mathcal{B}_r(t)$ with the integral of the boundary level $r(t)$ over the finite horizon time $[t_0,\,t_f]$ (resulting in a conservative under-approximation of $\mathcal{B}_r(t)$), the overall optimization problem can be stated as:
\begin{align}
    \underset{V(\bar{x},t),r(t),\bar{u}(\bar{y},t)}{\textrm{max}} \quad & \int_{t_0}^{t_f} r(t) \label{eq:opt}\\
    \textrm{s.t.} \quad & V(\bar{x},t)=r(t) \,\Rightarrow\, \dot{V}(\bar{x},\bar{u},w,t)<\dot{r}(t), \nonumber \\
    & \qquad \qquad \qquad \quad \forall t\in[t_0,t_f], \, \forall w\in \mathcal{W} \nonumber \\
    & u_{min} \leq \bar{u}(t) \leq  u_{max}, \; \forall t\in[t_0,t_f] \nonumber \\
    & \mathcal{B}_r(t_f) = \mathcal{X}_f\textrm{.} \nonumber
\end{align}

The initial Lyapunov function candidate is obtained using the time-varying LQR control design \cite{Farzan19} as $V_0(\bar{x},t)=\bar{x}^T(t)S(t)\bar{x}(t)$, where $S(t)$ is the solution to the differential Riccati equation $\dot{S}(t)=-(A^{T}S+SA-SBR^{-1}B^{T}S+Q)$,
with $A(t)$ and $B(t)$ as the Jacobian linearization of the original nonlinear system about the nominal trajectory, $Q=Q^T\geq0$, $Q_f=Q_f^T\geq 0$ and $R=R^T>0$ as the LQR cost matrices, and $S(t_f) = Q_f$. 

To include the time-varying Lyapunov function in the optimization decision variables, we decompose $V(\bar{x},t)$ as

\noindent \small
\begin{equation}
    V(\bar{x},t) = V_0(\bar{x},t)+ \bar{x}^TP(t)\bar{x}\textrm{,} \quad P(t)\geq 0\textrm{,}
\end{equation} \normalsize

\noindent where $V_0$ is the initial Lyapunov function described above, and $P(t)$ is a time-varying positive-semidefinite matrix with a fixed scale to be used as a decision variable. With the proposed decomposition, the time derivative $\dot{V}$ has the form:

\noindent \small
\begin{align}
    \dot{V}(\bar{x},\bar{u},w,t)= & \frac{\partial V_0(\bar{x},t)}{\partial \bar{x}}f_{cl}(\bar{x},\bar{u},w,t)+\frac{\partial V_0(\bar{x},t)}{\partial t} \\
    & +2\bar{x}^TP(t)f_{cl}(\bar{x},\bar{u},w,t)+\bar{x}^T\dot{P}(t)\bar{x}(t)\textrm{.} \nonumber
\end{align} \normalsize

\noindent To determine the goal set $\mathcal{B}_r(t_f)$, we use $V_0(\bar{x},t_f) = \bar{x}^TS(t_f)\bar{x}$, $P(t_f){=}0$ and $r(t_f){=}1$, with $S(t_f){=}Q_f$ as a $6\times6$ diagonal matrix which its diagonal elements determine the desired boundary on the final states, representing the set $\mathcal{X}_f$.\looseness=-1

\subsection{Sum-of-Squares Optimization Programs}
The polynomial S-procedure \cite{VanBoy96} and sum-of-squares relaxation technique \cite{Parrilo2003} for polynomial nonnegativity are used to express (\ref{eq:opt}) as a non-convex optimization in the form of semidefinite programs:

\noindent \small \vspace{-10pt}
\begin{subequations}
\begin{align}
    & \underset{P,r,\bar{u},L,L_{u,\{1,2\}},L_{w,\{1,2\}},L_{t,\{1,2,3\}}}{\textrm{max}} \quad \int_{t_0}^{t_f} r(t) \label{eq:opt-sos-a} \\
    \textrm{s.t.} \quad  &\dot{r}(t)-\dot{V}(\bar{x},\bar{u},w,t)-L\big(V(\bar{x},t)-r(t)\big)-L_{w,1}(w-w_{lb}) \nonumber \\
    & \qquad -L_{w,2}(w_{ub}-w)-L_{t,1}(t-t_0)(t_f-t) \geq 0 \label{eq:opt-sos-b}\\
    & \bar{u}(t)-u_{min}+L_{u,1}(V(\bar{x},t)-r(t)) \nonumber \\
    & \qquad \qquad \qquad \qquad \qquad -L_{t,2}(t-t_0)(t_f-t) \geq 0 \label{eq:opt-sos-c}\\
    & u_{max}-\bar{u}(t)+L_{u,2}(V(\bar{x},t)-r(t)) \nonumber \\
    & \qquad \qquad \qquad \qquad \qquad -L_{t,3}(t-t_0)(t_f-t) \geq 0 \label{eq:opt-sos-d}\\
    & L_{w,\{1,2\}},\,L_{u,\{1,2\}},L_{t,\{1,2,3\}}\geq 0
    \label{eq:opt-sos-e} \\
    & P(t)\geq0, \; P(t_f)=0, \; r(t)>0, \; r(t_f)=1\textrm{.} \label{eq:opt-sos-f}
\end{align}
\end{subequations}\normalsize

\noindent The decision variables consists of the Lyapunov function $V(\bar{x},t)$ (through $P(t)$), the boundary level $r(t)$, the control law $\bar{u}(\bar{y},t)$, and the set of Lagrange multipliers $L(\bar{x},w,t)$, $L_{u,\{1,2\}}(\bar{x},t)$, $L_{w,\{1,2\}}(\bar{x},w,t)$ and $L_{t,\{1,2,3\}}(\bar{x},w,t)$ as S-procedure polynomial certificates. To reformulate the optimization as a convex problem, the problem can be solved by an iterative, three-way search between the two bilinear pairs involving the decision variables $(L(\bar{x},w,t),\,r(t))$, and $(\bar{u}(t),\,V(\bar{x},t))$, as stated in optimizations (\ref{eq:step1}) to (\ref{eq:step3}).

\noindent i) Fix the Lyapunov function $V(\bar{x},t)$ and the boundary level $r(t)$, and introduce the slack variable $\gamma$,

\noindent \small
\begin{equation}
    \underset{\gamma,\bar{u},L,L_{u,\{1,2\}},L_{w,\{1,2\}},L_{t,\{1,2,3\}}}{\textrm{min}} \quad \gamma \label{eq:step1}
\end{equation}
\begin{align}
     & \qquad \textrm{s.t.} \quad  \gamma-\Big[\dot{V}(\bar{x},\bar{u},w,t)-\dot{r}(t)+L\big(V(\bar{x},t)-r(t)\big) \nonumber \\ 
    & \hspace{-0.1in} +L_{w,1}(w-w_{lb})+L_{w,2}(w_{ub}-w)+L_{t,1}(t-t_0)(t_f-t)\Big] \geq 0 \nonumber \\
    & \qquad \qquad (\ref{eq:opt-sos-c}), \; (\ref{eq:opt-sos-d}), \; (\ref{eq:opt-sos-e}) \nonumber
\end{align}
\normalsize

\noindent ii) Fix the Lagrange multiplier $L(\bar{x},w,t)$ and the Lyapunov function $V(\bar{x},t)$,

\noindent \small
\begin{align}
    & \underset{r,\bar{u},L_{u,\{1,2\}},L_{w,\{1,2\}},L_{t,\{1,2,3\}}}{\textrm{max}}  \quad  \int_{t_0}^{t_f} r(t) \label{eq:step2} \\
    & \qquad \qquad \quad \; \textrm{s.t.} \qquad \; (\ref{eq:opt-sos-b}), \; (\ref{eq:opt-sos-c}), \; (\ref{eq:opt-sos-d}), \; (\ref{eq:opt-sos-e}), \; (\ref{eq:opt-sos-f}) \nonumber
\end{align} \normalsize

\noindent iii) Fix the control law $\bar{u}(\bar{y},t)$ and the Lagrange multipliers $L(\bar{x},w,t)$ and $L_{u,\{1,2\}}(\bar{x},t)$,

\noindent \small
\begin{align}
    \underset{r,P,L_{w,\{1,2\}},L_{t,\{1,2,3\}}}{\textrm{max}} \quad & \int_{t_0}^{t_f} r(t) \label{eq:step3} \\
    \textrm{s.t.} \qquad \;  & \; (\ref{eq:opt-sos-b}), \; (\ref{eq:opt-sos-c}), \; (\ref{eq:opt-sos-d}), \; (\ref{eq:opt-sos-e}), \; (\ref{eq:opt-sos-f}) \nonumber
\end{align} \normalsize
The three-step optimization search has converged when no more improvement is observed in $\int r(t)$.

It is important to note that to exploit the use of sum-of-squares technique, the system dynamics $f_{cl}$ and the control law $\bar{u}(t)$ are restricted to be polynomials. To that end, the nonlinear equations of motion in (\ref{eq:eom}) are converted to polynomial dynamics by Taylor expansion around the nominal trajectory. Moreover, for practical computation, the optimization problems in equations (\ref{eq:step1}) to (\ref{eq:step3}) are implemented by time-sampling \cite{MajTed17}, where the constraints are checked only at sample times $t_i\in[t_0,\,t_f],\;i\in\{1,\ldots,N\}$. Since brachiation maneuvers are about $0.8$ seconds long, we use $N=40$ to achieve $20$ ms sample intervals, which results in a close approximation for our application.

\subsection{Library of Trajectories and SOS-based Controllers} \label{subsec:hybrid}
The parametric model uncertainty in the system results in restricted inner-approximation of the backward reachable sets for the optimal trajectories, as will be shown in Section \ref{sec:results}.
Inspired by the 
funnel libraries \cite{MajTed17} algorithm, we can employ several optimal trajectories and their associated SOS-based controllers to enclose a larger set of initial configurations and state-space by the controller. The optimal trajectories are generated from different initial configurations on the cable, while all drive the robot to the same set of desired final configurations $\mathcal{X}_f$. By using a sufficient number of trajectories covering the entire range of initial conditions on the cable, we can create a feedback motion planning platform to robustly control the brachiating robot on the flexible cable from all possible initial configurations.


\section{SIMULATION RESULTS AND HARDWARE EXPERIMENTS} \label{sec:results}
The physical parameters of the robot-cable system used for the SOS synthesis/verification and in the simulations and hardware experiments are summarized in Table \ref{tab:physical-parameters}.
It is assumed that brachiating maneuvers are performed on an 8 meter flexible cable.
For the proposed cable dynamic model, the equivalent spring-damper parameters to such cable (derived by the system ID procedure) are listed in Table \ref{tab:sys-id}.  The uncertainty in the cable dynamics is taken into account in our computations by considering $20\%$ parametric model uncertainty in the stiffness of the three springs. That is, the uncertainty parameter in (\ref{eq:stiffness}) is set to $w=0.2$, resulting in $k_i\in[0.8k_{0_i},\,1.2k_{0_i}]$.

Using the parametric trajectory optimization method presented in \cite{Farzan18}, an open-loop optimal reference trajectory for a single brachiating maneuver over the flexible cable was generated for the initial and final states of $[-45^{\circ},-90^{\circ},1.84\,\textrm{m},0,0,0]$ and $[45^{\circ},90^{\circ},1.9\,\textrm{m},120\,\textrm{(deg/s)},120\,\textrm{(deg/s)},0]$ respectively, associated to $[\theta_1,\,\theta_2,\,z_g,\,\dot{\theta}_1,\,\dot{\theta}_2,\,\dot{z}_g]$. The finite time horizon is set to $t\in[0,\,0.7]$ seconds, according to the reference trajectory. The control input is constrained to $u\in[-5,5]$ Nm based on the torque limits of the actuator installed on the robot hardware prototype. For initialization of and comparison to the SOS-based controller, a TVLQR feedback controller was designed for the same robot as presented in \cite{Farzan19}.\looseness=-1

The performance of the SOS-based output feedback controller for the robot with parametric model uncertainty is demonstrated experimentally, and is evaluated in terms of three criteria: i) the size of the approximated backward reachable set for the SOS-based controller compared to the TVLQR controller, ii) the robot performance under the controller starting from initial conditions different than the optimal nominal trajectory and with cable stiffness different than the nominal value, iii) the robot performance under the controller when executing multiple sequential swings to traverse the entire length of the cable.

The optimizations and simulations in this section are computed on a workstation with a 3.0 GHz Intel Core i7 processor and 32 GB of RAM. The SOS optimization programs are expressed as SDP problems using the YALMIP toolbox \cite{Lofberg2004}, and solved by the MOSEK optimization toolbox \cite{mosek} for MATLAB. 

\begin{table}[t]
\caption{\label{tab:physical-parameters}Physical parameters of the robot and the cable}
\noindent \begin{centering}
\setlength\tabcolsep{3pt}
\begin{tabular}{|c|c|}
\hline 
{\textit{\footnotesize{}Parameter}} & \textit{\footnotesize{}Value}\tabularnewline
\hline 
{\footnotesize{}Main body center of mass} & {\footnotesize{}$m_{0}=1.247\,\textrm{kg}$}\tabularnewline
\hline 
{\footnotesize{}Link 1 and 2 center of mass} & {\footnotesize{}$m=m_{1}=m_{2}=0.794\,\textrm{kg}$}\tabularnewline
\hline 
{\footnotesize{}Link 1 and 2 length} & {\footnotesize{}$l=l_{1}=l_{2}=0.35\,\textrm{m}$}\tabularnewline
\hline 
{\footnotesize{}Link 1 \& 2 center of mass location} & {\footnotesize{}$d_{1}=0.15\,\textrm{m}$, $d_{2}=0.2\,\textrm{m}$} \tabularnewline
\hline 
{\footnotesize{}Link 1 and 2 moment of inertia } & {\footnotesize{}$I_{1}=I_{2}=0.0088\,\textrm{kg\,m}^2$}\tabularnewline
\hline 
{\footnotesize{}Cable length and linear mass} & {\footnotesize{}$l_{c}=8\,\textrm{m}$, $\;m_{c}=0.25\,\textrm{kg/m}$} \tabularnewline
\hline 
{\footnotesize{}Cable stiffness and damping} & {\footnotesize{}$k_{c}=785400\,\textrm{N/m}$, $\;b_{c}=4\,\textrm{Ns/m}$} \tabularnewline
\hline 
\end{tabular}
\par\end{centering}{\small \par}
\end{table}

\begin{figure}[t]
    \vspace{-8pt}
    \begin{minipage}{0.25\textwidth}
    {\includegraphics[trim={60bp 2bp 102bp 28bp}, clip,width=\linewidth]{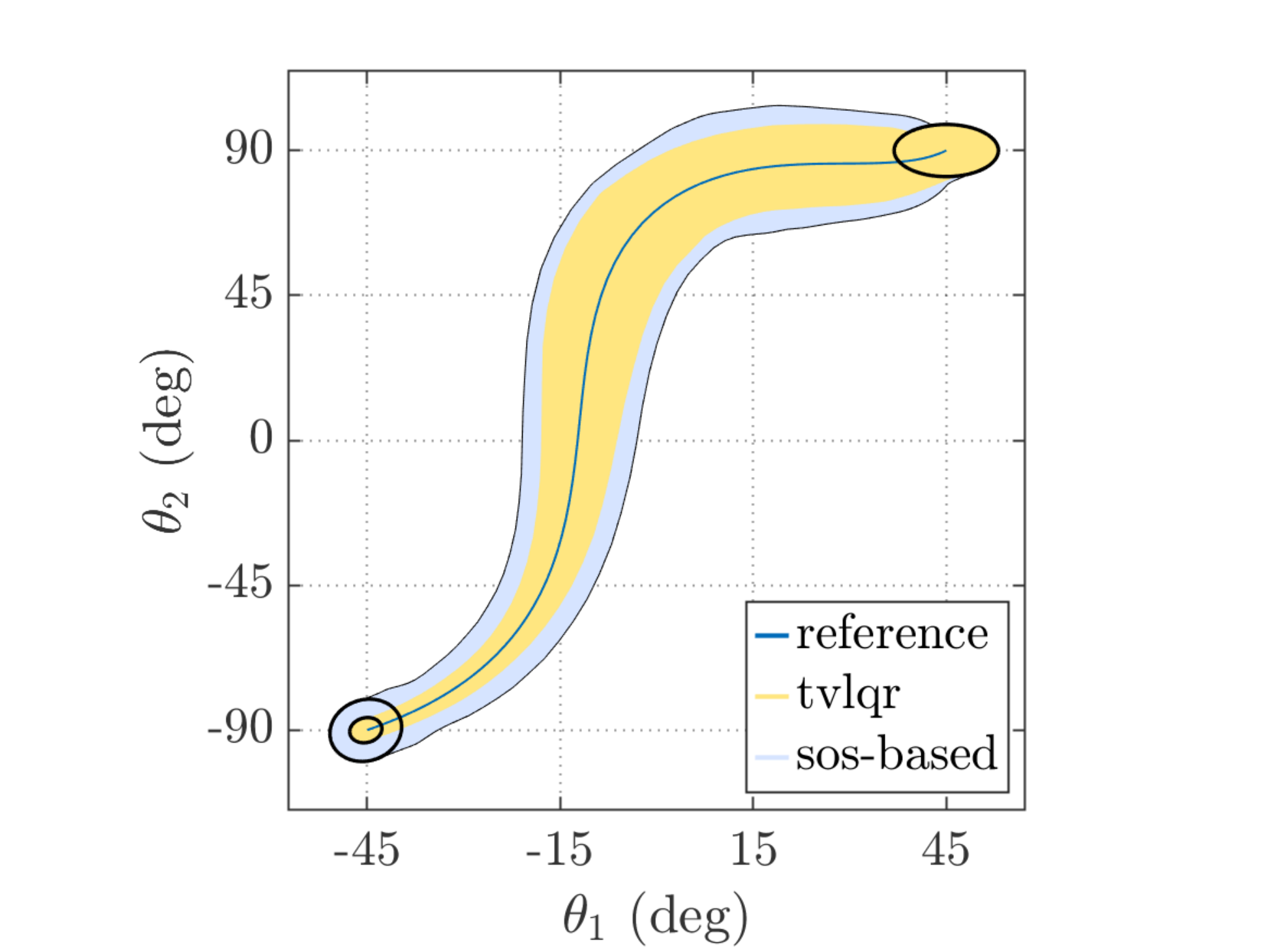}}
    \end{minipage}%
    \begin{minipage}{0.23\textwidth}
    {\includegraphics[trim={0bp 50bp 45bp 90bp}, clip,width=\linewidth]{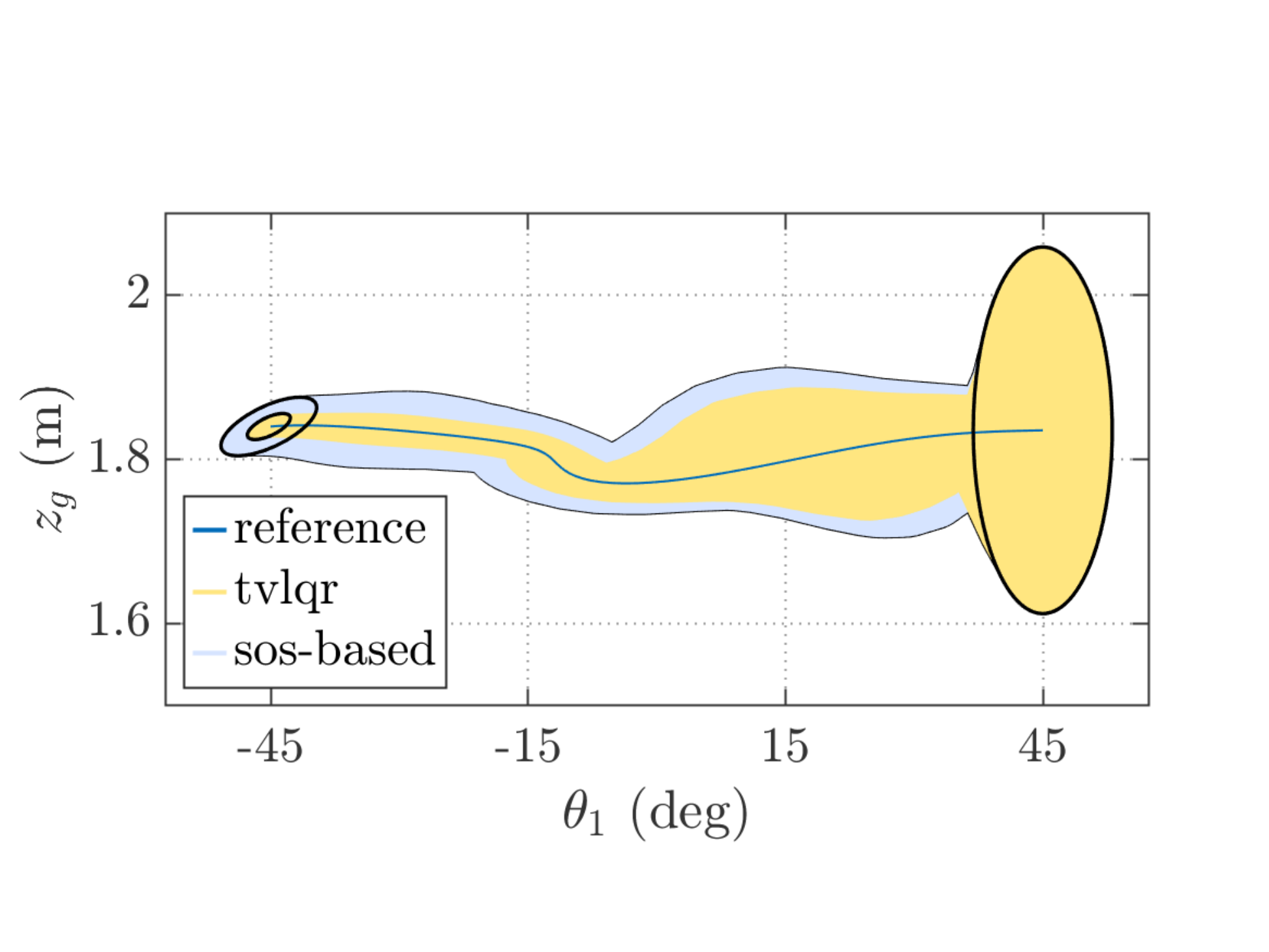}\\
    \includegraphics[trim={0bp 50bp 45bp 90bp}, clip,width=\linewidth]{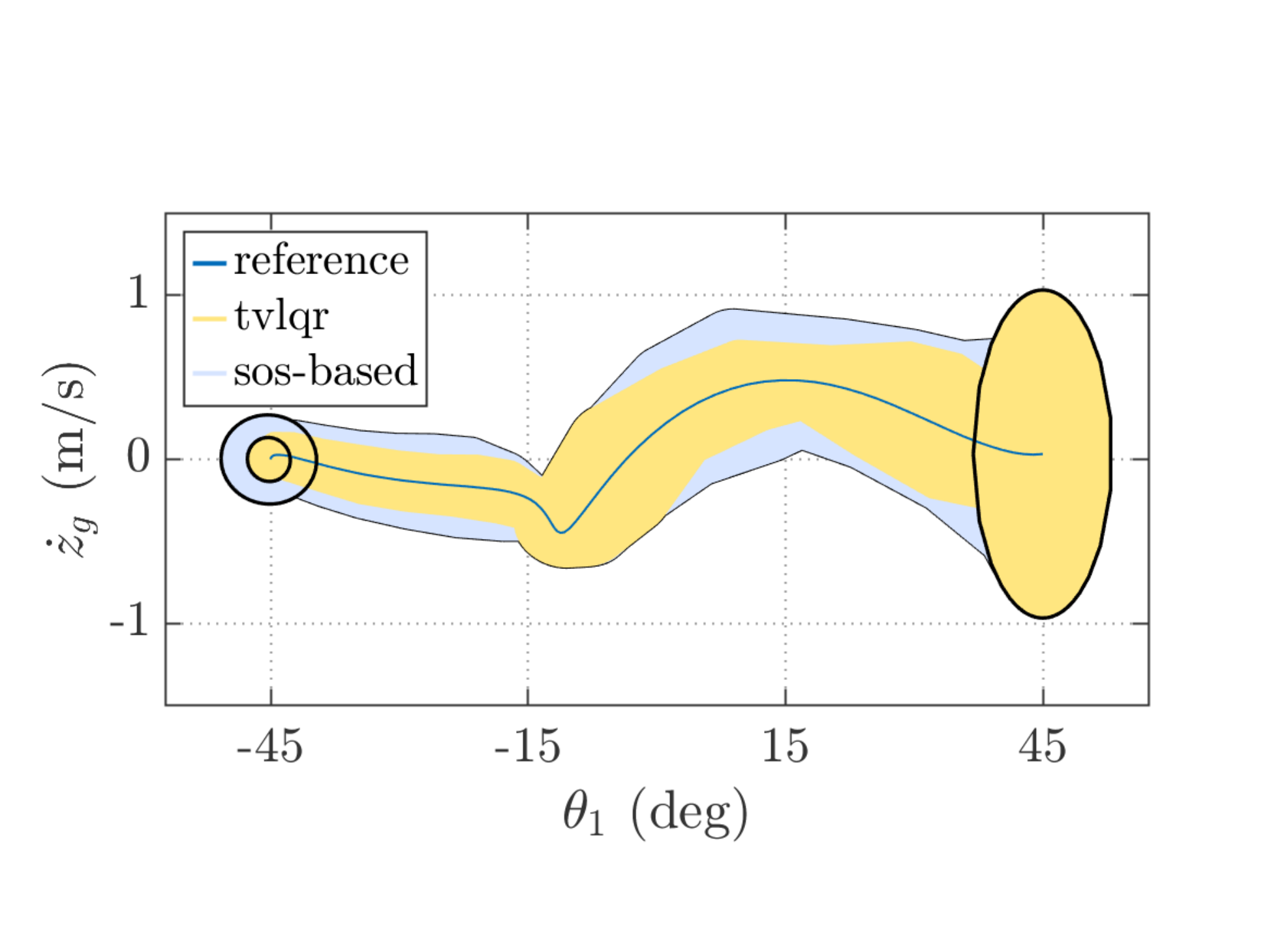}}
    \end{minipage} \\
    \begin{minipage}{.25\textwidth}
    \centering \scriptsize{(a)}
    \end{minipage}%
    \begin{minipage}{0.23\textwidth}
    \centering \scriptsize{(b), (c)}
    \end{minipage} \\
    \begin{minipage}{.24\textwidth}
    {\includegraphics[trim={0bp 50bp 45bp 90bp}, clip,width=\linewidth]{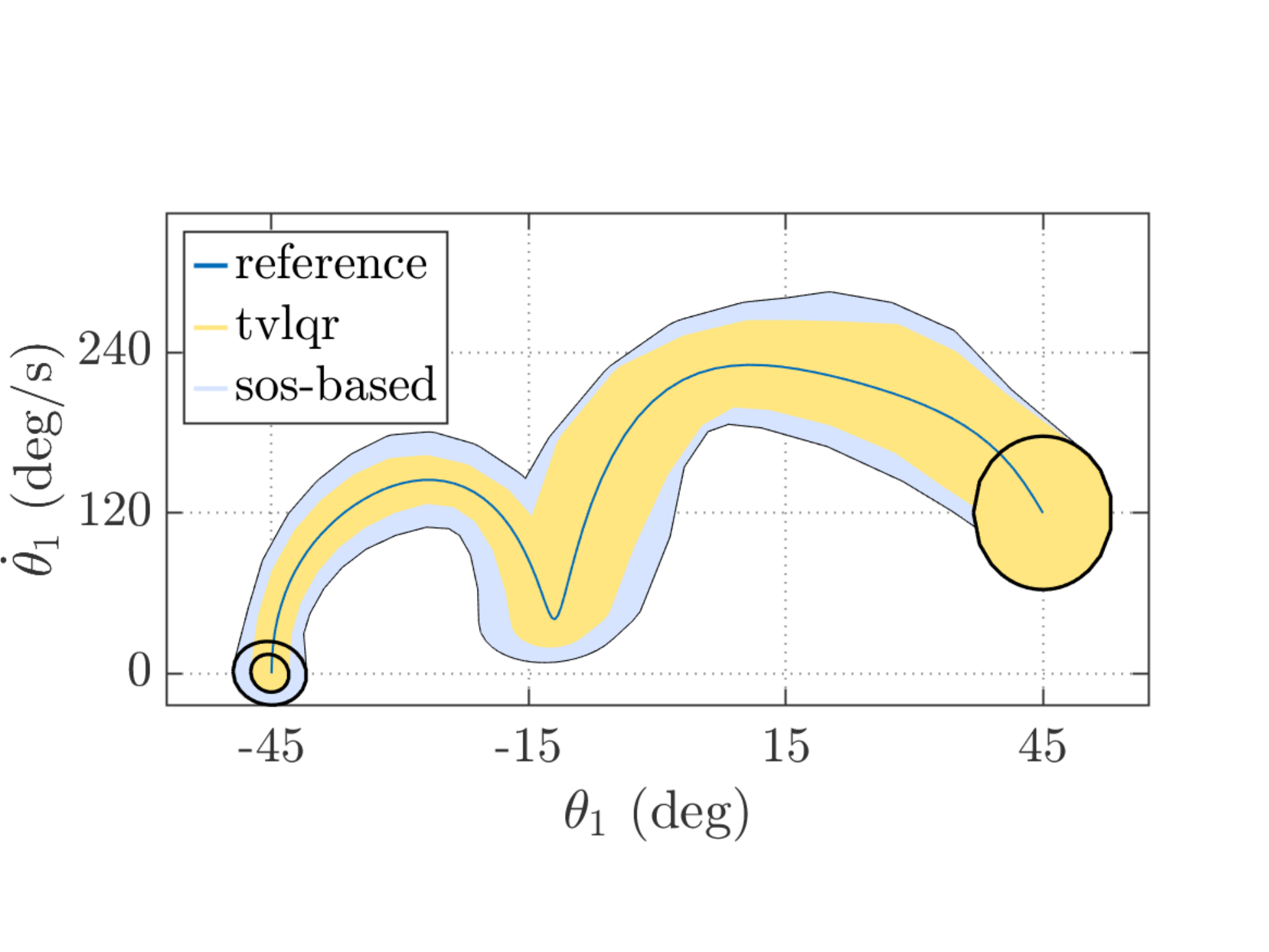}}
    \end{minipage}%
    \begin{minipage}{.24\textwidth}
    {\includegraphics[trim={0bp 50bp 45bp 90bp}, clip,width=\linewidth]{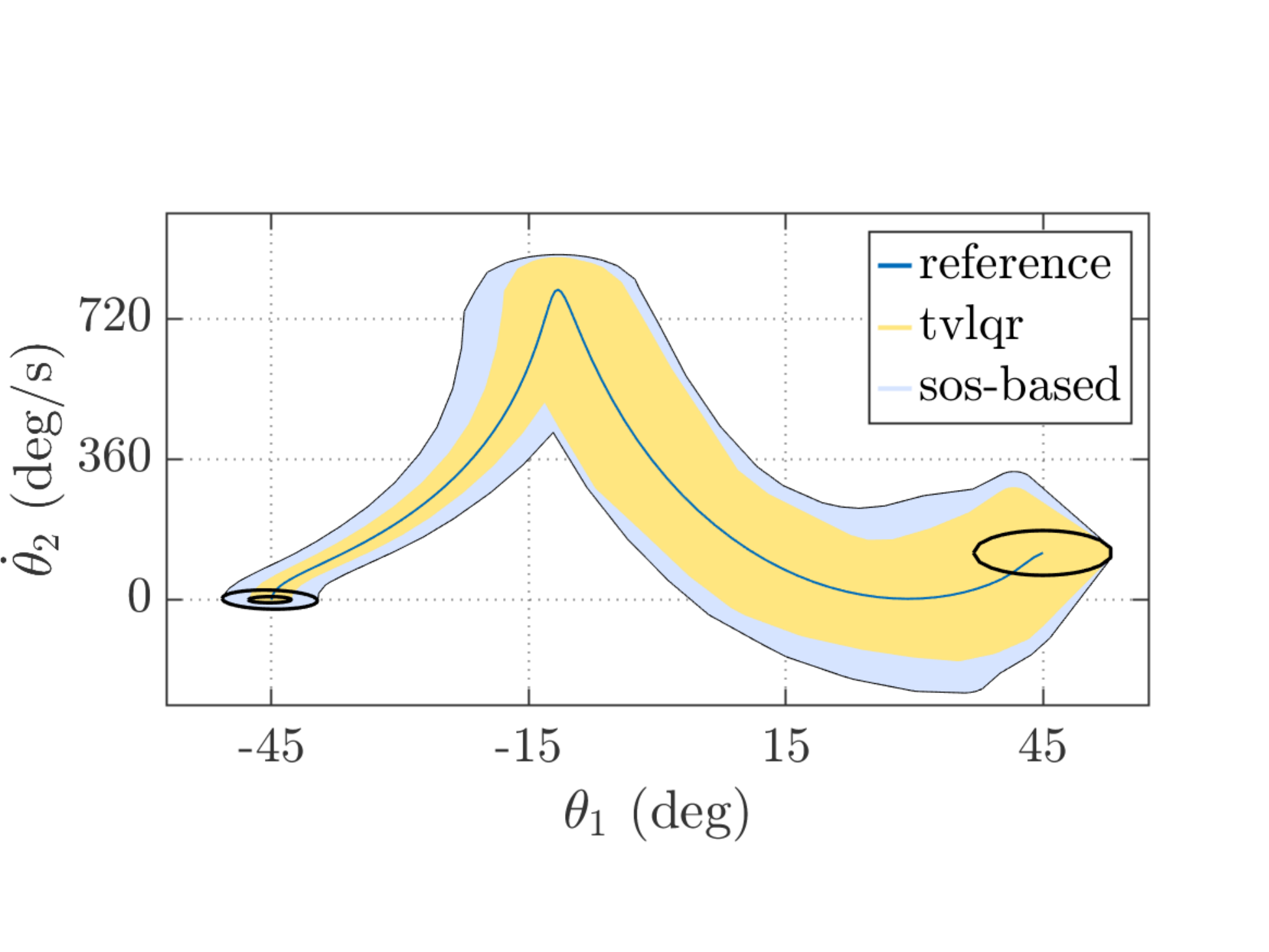}}
    \end{minipage} \\
    \begin{minipage}{0.24\textwidth}
    \centering \scriptsize{(d)}
    \end{minipage}%
    \begin{minipage}{0.24\textwidth}
    \centering \scriptsize{(e)}
    \end{minipage}
    \vspace{-5pt}
\caption{\label{fig:invset}Comparison of the inner-approximation of the robust backward reachable set for the SOS-based and the TVLQR controllers, projection of $\theta_1$ vs. (a) $\theta_2$, (b) $z_g$, (c) $\dot{z}_g$, (d) $\dot{\theta}_1$, (e) $\dot{\theta}_2$.}
\end{figure}

\subsection{Robust Control Synthesis and Verification Results} \label{subsec:results-set}
The iterative optimization algorithm described in (\ref{eq:step1}) to (\ref{eq:step3}) was carried out for the brachiating robot system detailed above. We used polynomials of degree 4 for the Lagrange multipliers $L$, $L_u$, $L_w$ and $L_t$, while the degree of the controller polynomial $\bar{u}$ is set to 1. The computing time required for the offline optimization convergence was approximately 4 hours.
The long time required for convergence is not an issue for practical implementation of the controller, as the resulting feedback control policy $\bar{u}(\bar{y},t)$ (represented by time-varying gains on measurable states) will be hard-coded into the robot.\looseness=-1

To visualize the resulting robust backward reachable set, we project its 2-dimensional subspaces (out of the full 6-dimensional state-space) on 2D plots. 
Fig. \ref{fig:invset} shows the projections of each state vs. $\theta_1$, and compares
the inner-approximation of the robust backward reachable sets for both the SOS-based controller and the time-varying LQR controller. As shown on the plots, the resulting invariant sets for the SOS-based controller cover a larger part of the state-space compared to TVLQR. The inner-approximation of the backward reachable set for the TVLQR controller is computed by solving the SOS program in (\ref{eq:opt-sos-a}) without including the controller $\bar{u}$ in the optimization decision variables, eliminating the need for the second step optimization in (\ref{eq:step2}).\looseness=-1

\begin{figure}[t]
\renewcommand{\arraystretch}{0.25}
\setlength\tabcolsep{1pt}
\begin{tabular}{ccc}
\includegraphics[trim={60bp 1bp 104bp 30bp}, clip,height=1.0in,width=0.33\columnwidth]{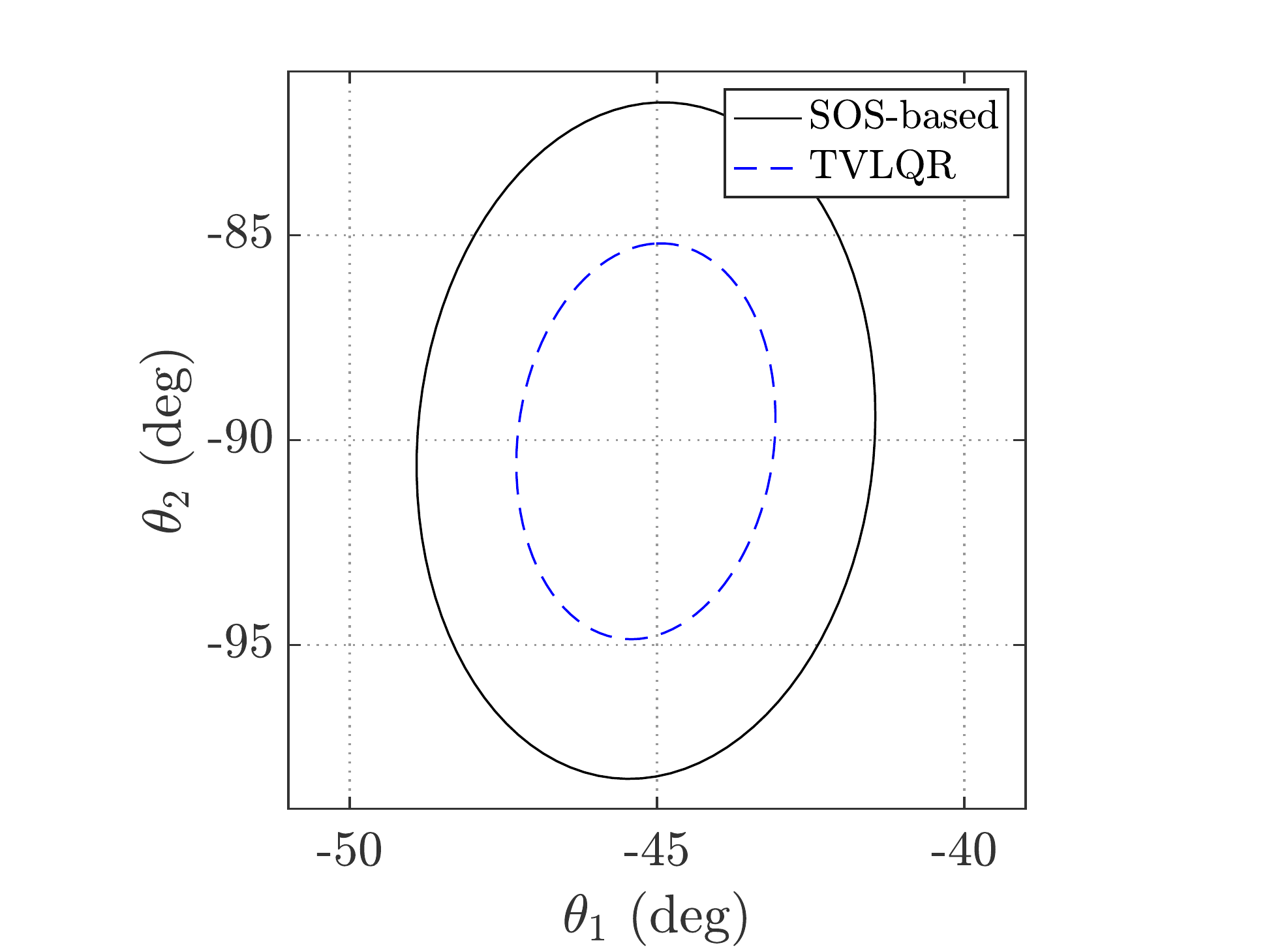} &
\includegraphics[trim={55bp 1bp 104bp 30bp}, clip,height=1.0in,width=0.33\columnwidth]{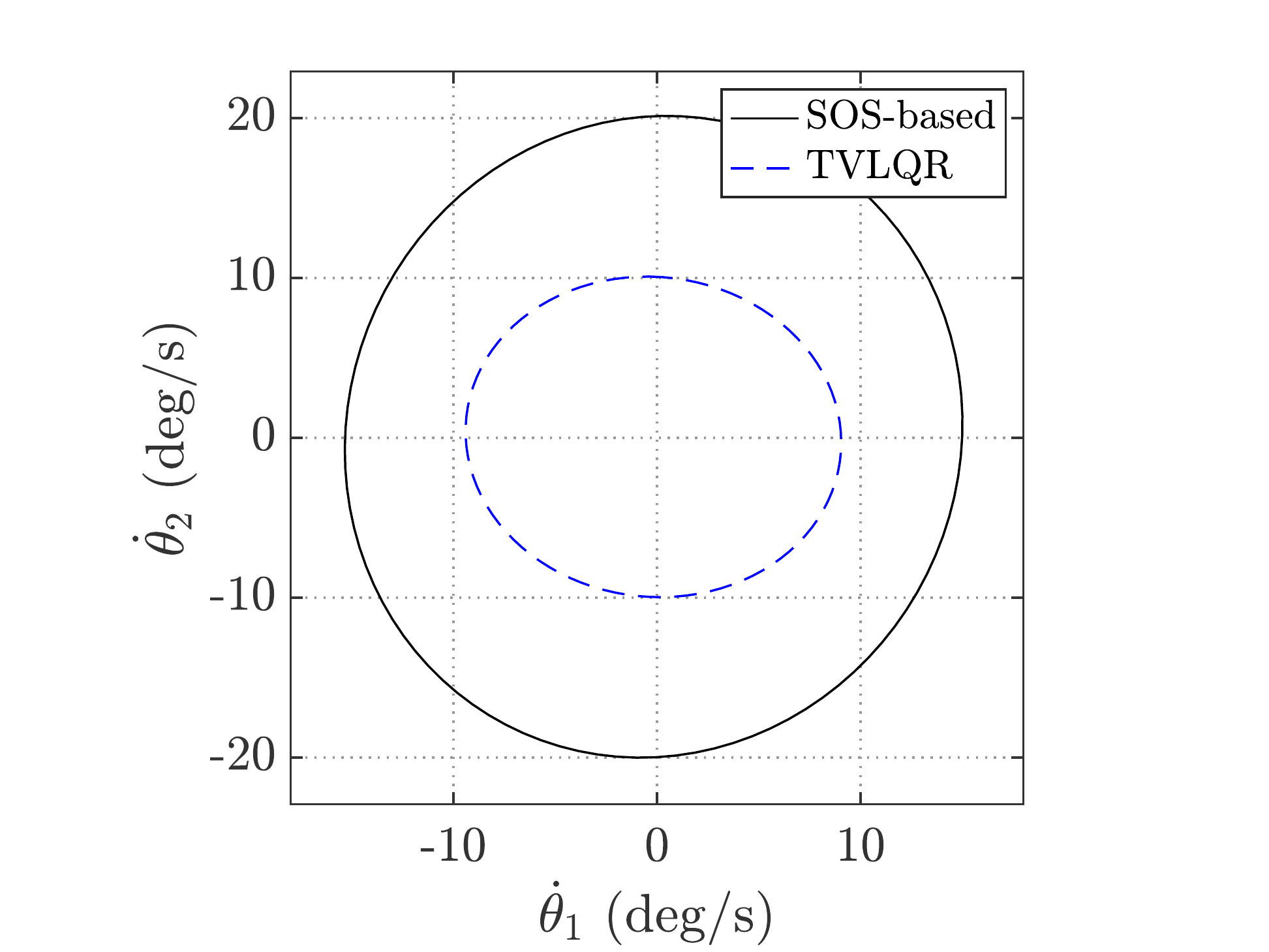} &
\includegraphics[trim={55bp 1bp 104bp 30bp}, clip,height=1.0in,width=0.33\columnwidth]{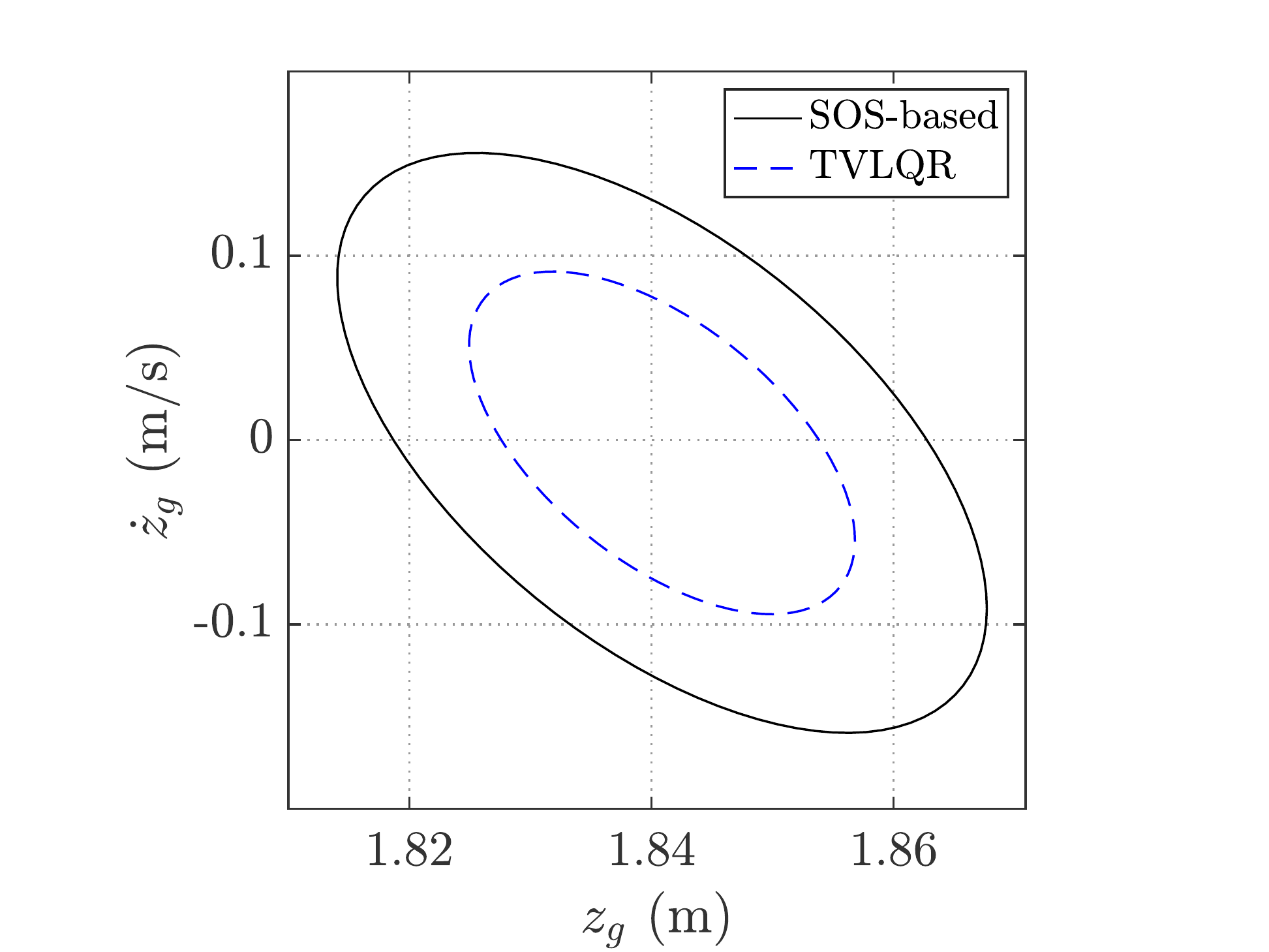}
\tabularnewline
{\scriptsize{}(a)} & {\scriptsize{}(b)} & {\scriptsize{}(c)}
\tabularnewline
\end{tabular}
\caption{\label{fig:invset-init} Comparison of the verified set of initial conditions $\mathcal{x}_0$ which are driven to the desired set $\mathcal{X}_f$ for the SOS-based and the TVLQR controllers, projection of (a) $\theta_1$ vs. $\theta_2$, (b) $\dot{\theta}_1$ vs. $\dot{\theta}_2$, and (c) $z_g$ vs. $\dot{z}_g$.}
\end{figure}

Furthermore, as depicted in Fig. \ref{fig:invset-init}, the verified set of initial conditions $\mathcal{X}_0$ which is driven to the desired set $\mathcal{X}_f$ by the SOS-based controller is larger in every dimension compared to the corresponding set for the TVLQR controller.

\subsection{Validation by Simulation Experiments}
\begin{figure}[t]
\vspace{-2pt}
\begin{minipage}{0.25\textwidth}
    {\includegraphics[trim={60bp 2bp 102bp 28bp}, clip,width=\linewidth]{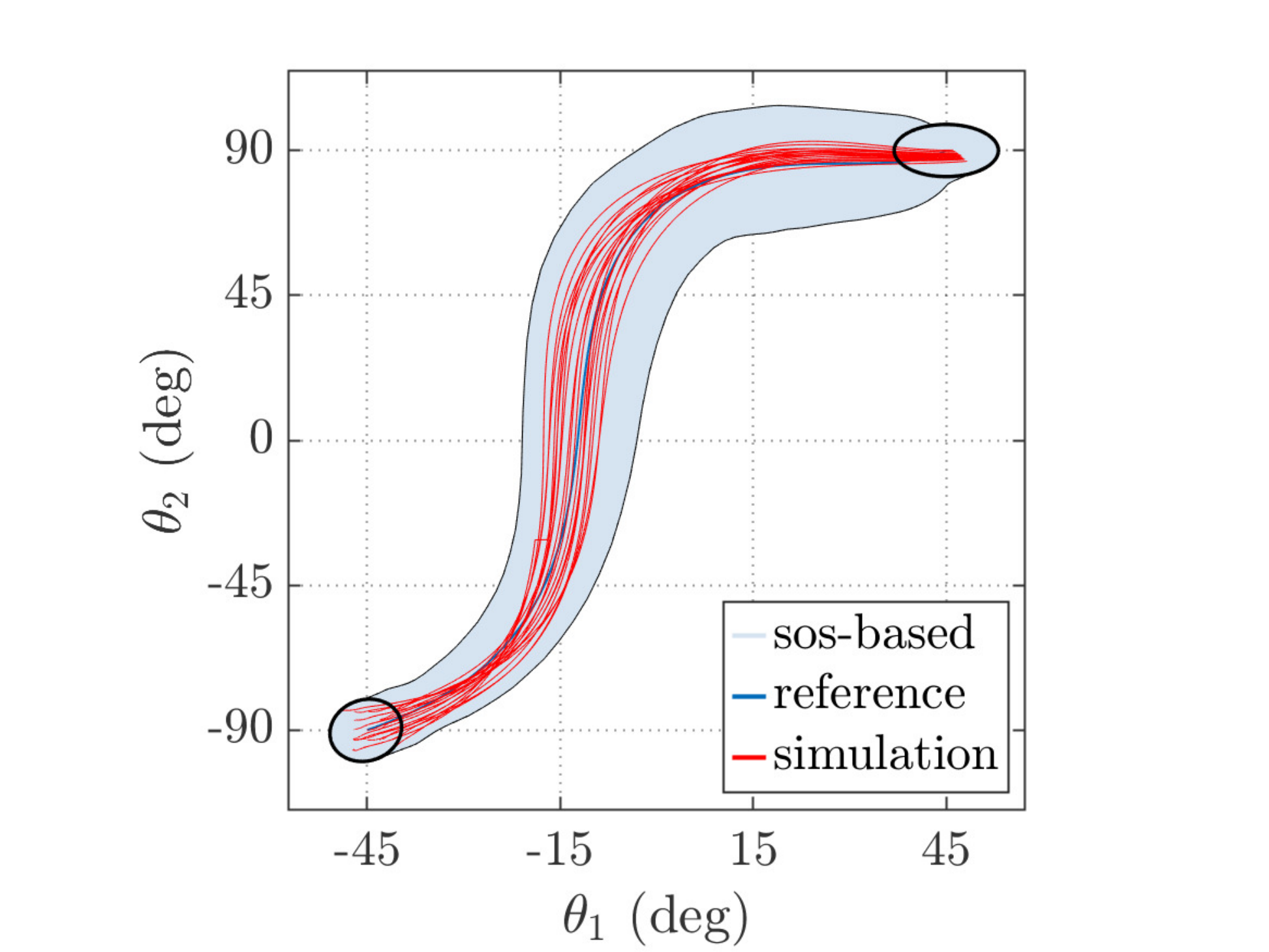}}
\end{minipage}%
\begin{minipage}{0.23\textwidth}
    \includegraphics[trim={0bp 50bp 45bp 90bp},clip,width=\linewidth]{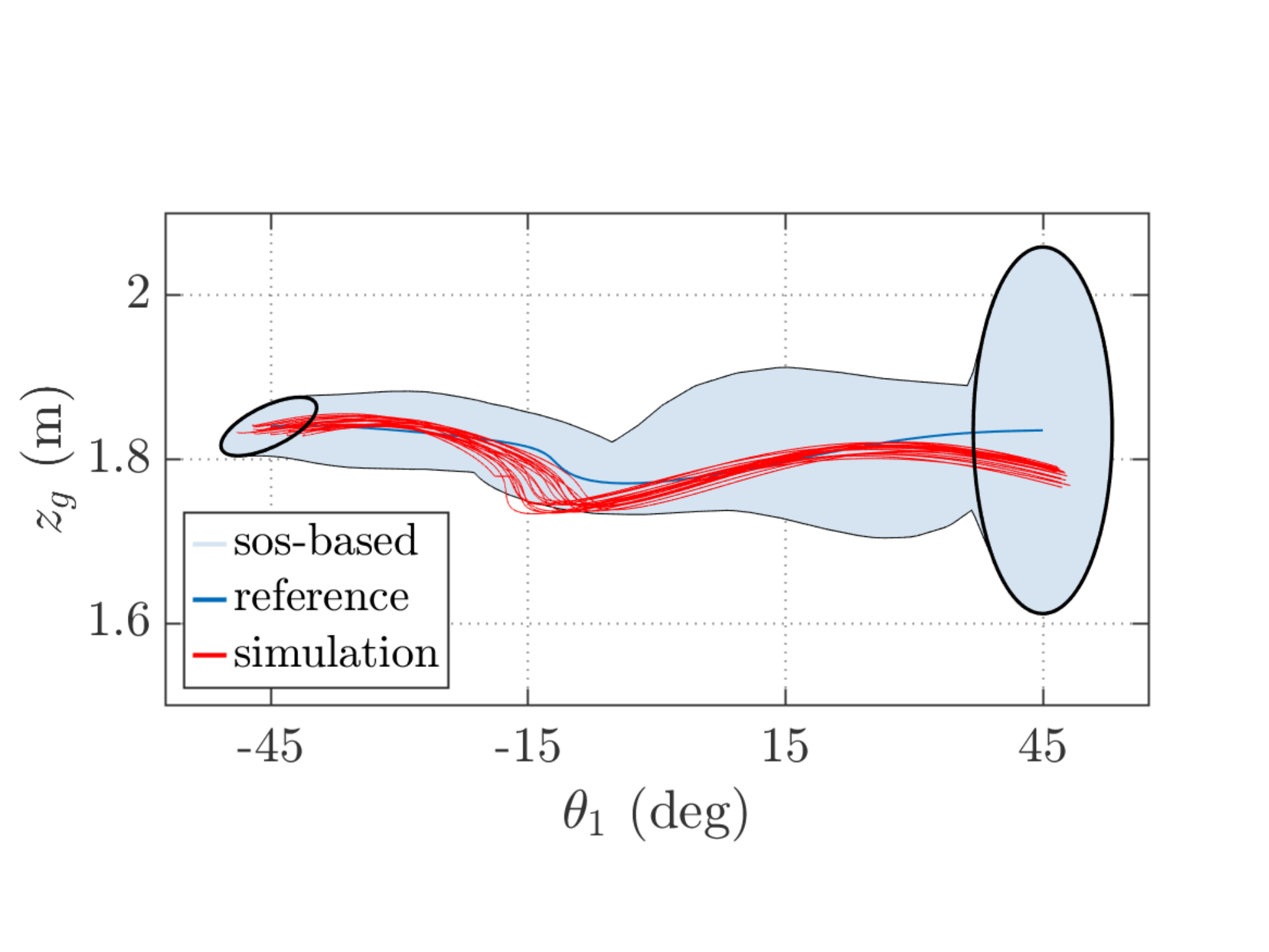} \\
    \includegraphics[trim={0bp 50bp 45bp 90bp},clip,width=\linewidth]{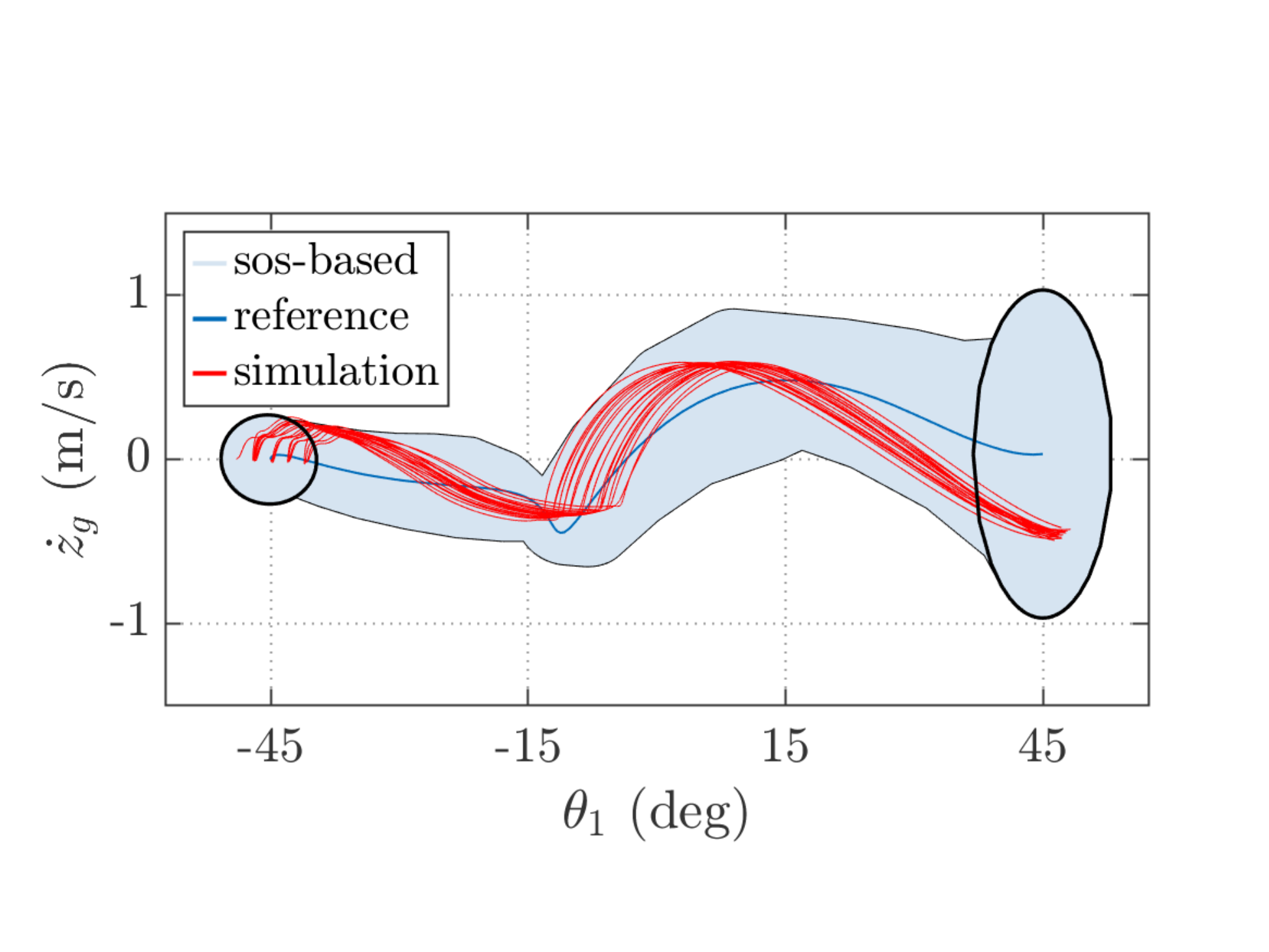}
\end{minipage} \\
\begin{minipage}{0.25\textwidth}
    \centering \scriptsize{(a)}
\end{minipage}%
\begin{minipage}{0.23\textwidth}
    \centering \scriptsize{(b) , (c)}
\end{minipage} \\
\begin{minipage}{0.24\textwidth}
\includegraphics[trim={0bp 50bp 45bp 90bp},clip,width=\linewidth]{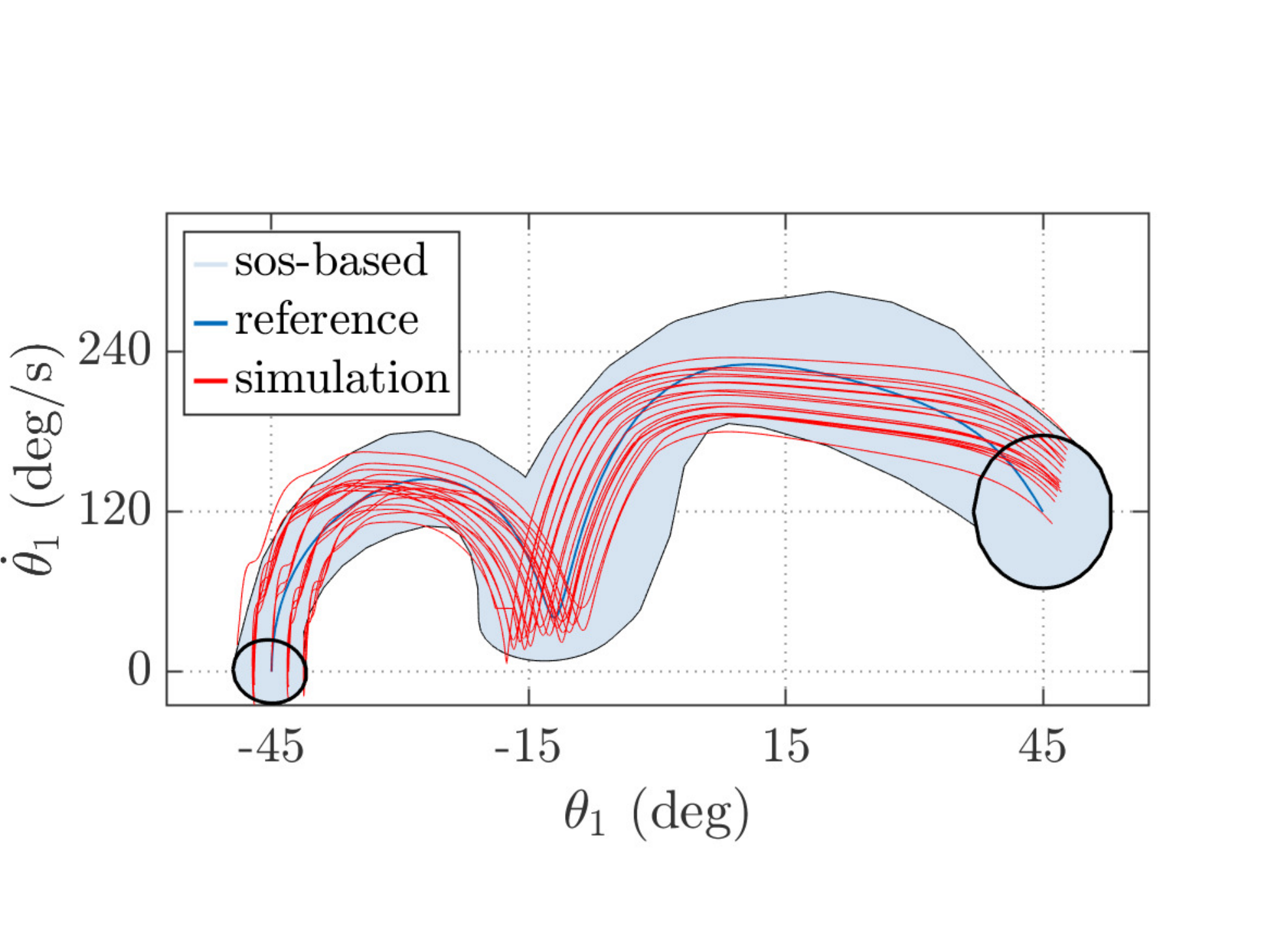}
\end{minipage}%
\begin{minipage}{0.24\textwidth}
\includegraphics[trim={0bp 50bp 45bp 90bp},clip,width=\linewidth]{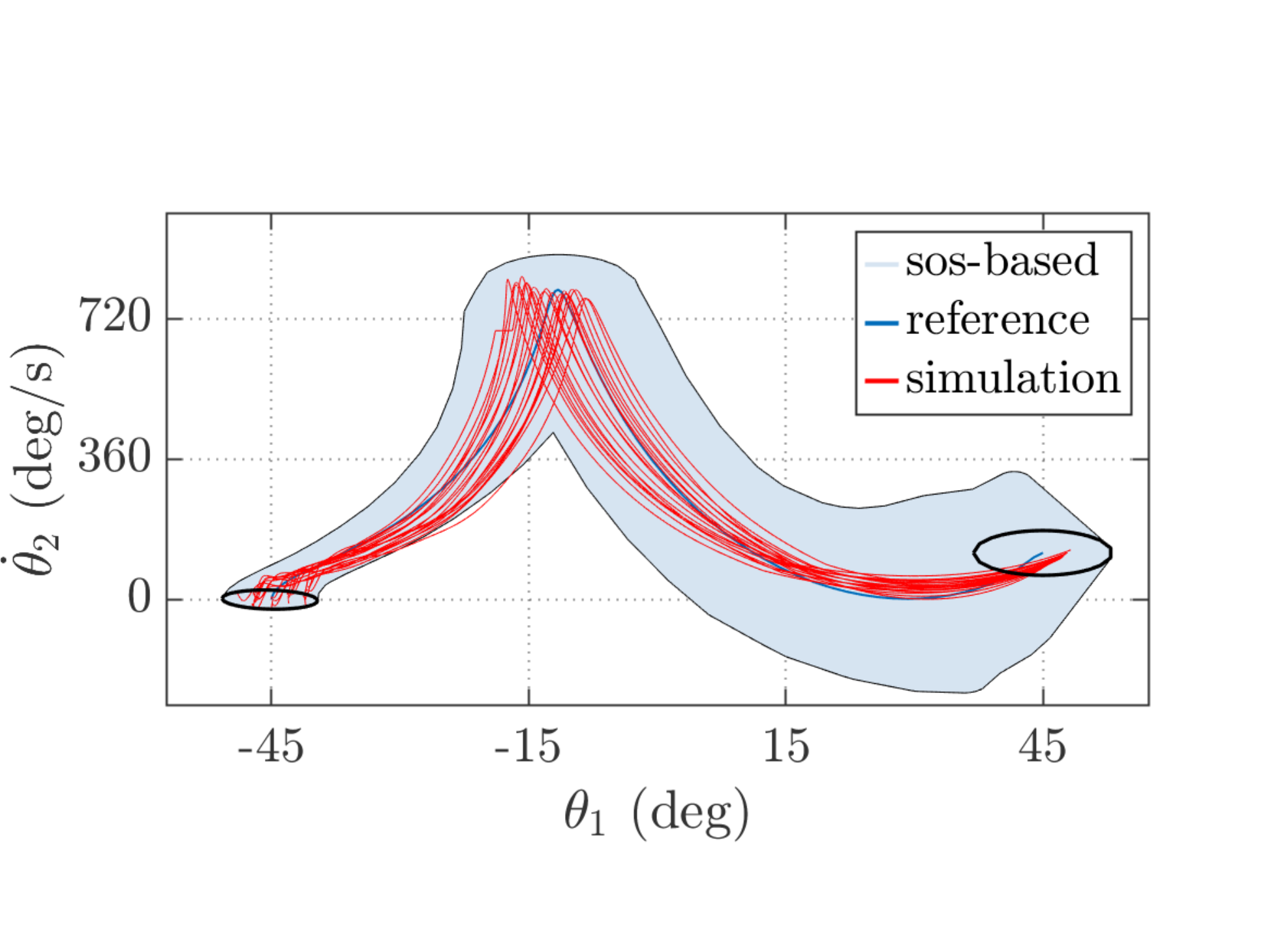}
\end{minipage} \\
\begin{minipage}{0.24\textwidth}
    \centering \scriptsize{(d)}
\end{minipage}%
\begin{minipage}{0.24\textwidth}
    \centering \scriptsize{(e)}
\end{minipage}
\vspace{-5pt}
\caption{\label{fig:invset-simulation} Simulated motion trajectories with the SOS-based controller (for brachiation on full-cable model with $20\%$ stiffness error starting from random initial configurations) plotted on top of the approximated backward reachable set. Projection of $\theta_1$ vs. (a) $\theta_2$, (b) $z_g$, (c) $\dot{z}_g$, (d) $\dot{\theta}_1$, (e) $\dot{\theta}_2$.}
\end{figure}

The performance of the robust SOS-based controller as well as the inner-approximation of its backward reachable set are validated by 20 simulation trials of the brachiating robot attached to the full-cable model. The stiffness of the cable is set to $20\%$ less than the nominal value. The robot starts from random initial conditions on the cable within the verified set of initial condition, and
the time-varying SOS-based controller is applied for the time horizon of the nominal trajectory. In Fig. \ref{fig:invset-simulation}, the resulting motion trajectories are plotted on top of the projections of the robust verified regions computed in Section \ref{subsec:results-set}. As can be seen on the plots, the resulting brachiating motion trajectories under the feedback controller lie within the verified backward reachable set for most of the experiments. Note that a few trajectories leave the verified region on $\dot{\theta}_1$ dimension, which could be explained by the conservative and inner-approximation formulation used to derive the invariant sets.

\begin{figure}[t]
\vspace{-10pt}
\begin{minipage}{0.245\textwidth}
{\begin{tikzpicture}\node[inner sep=0pt] at (0,0)
{\includegraphics[trim={3bp 1bp 35bp 20bp},clip,width=\linewidth]{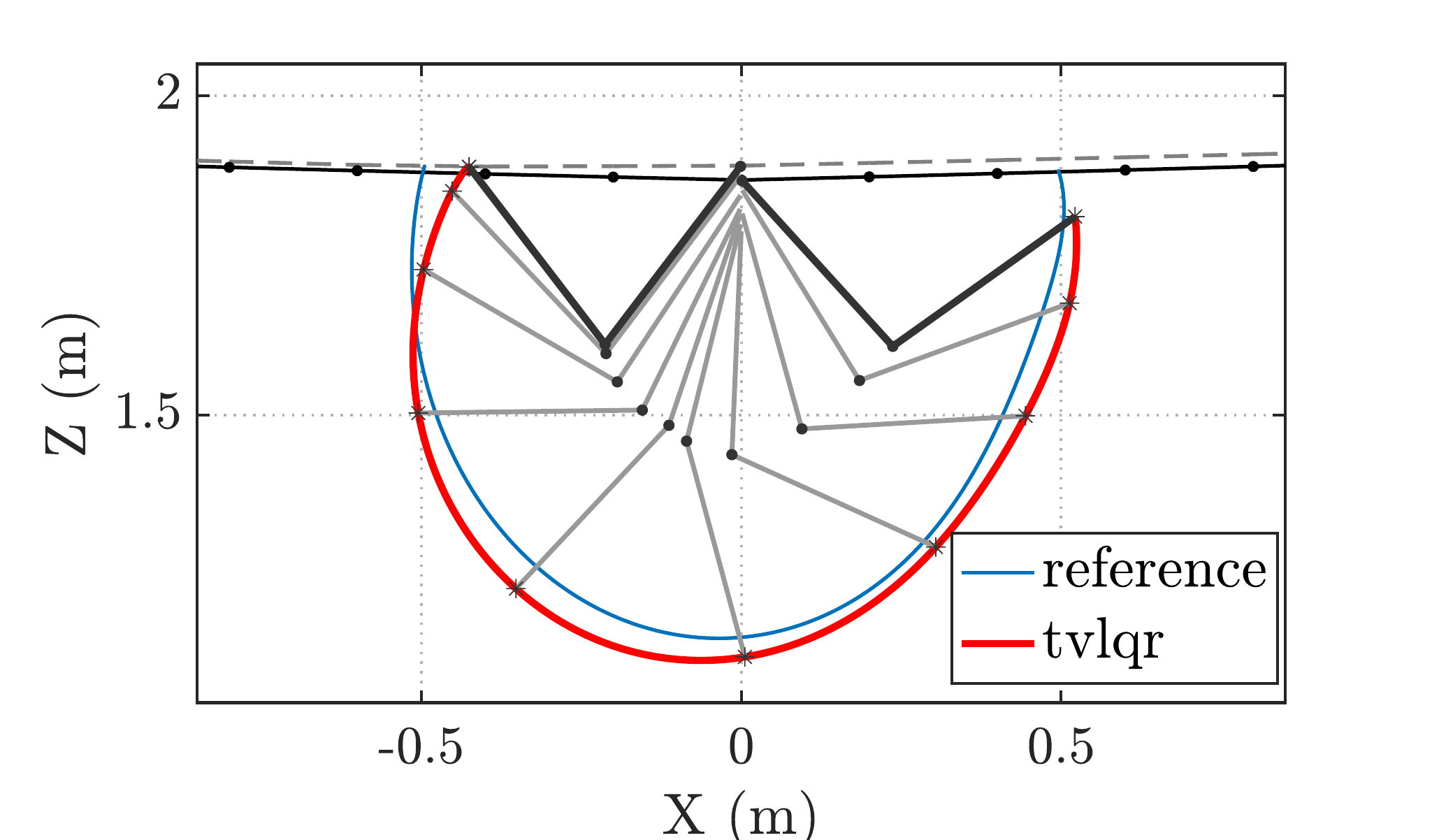}};
\draw[<-,>=stealth',semithick,red,dashed] (-1.0,-0.05) to [out=130,in=230] (-1.0,0.8);
\end{tikzpicture}}
\end{minipage}%
\begin{minipage}{0.245\textwidth}
{\begin{tikzpicture}\node[inner sep=0pt] at (0,0)
{\includegraphics[trim={3bp 1bp 35bp 20bp},clip,width=\linewidth]{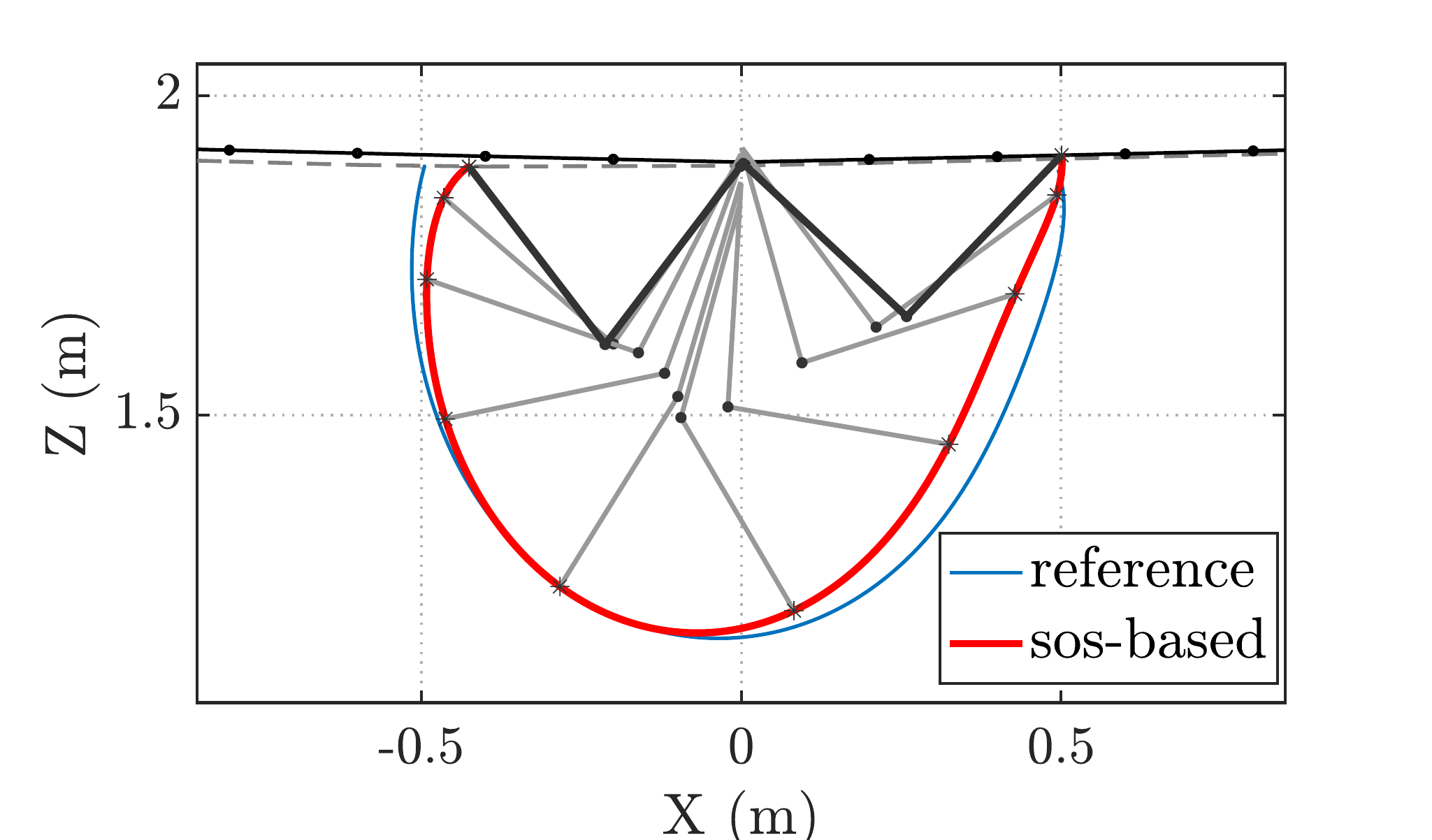}};
\draw[<-,>=stealth',semithick,red,dashed] (-1.0,-0.05) to [out=130,in=230] (-1.0,0.8);
\end{tikzpicture}}
\end{minipage} \\
\begin{minipage}{0.245\textwidth}
    \centering \scriptsize{(a)}
\end{minipage}%
\begin{minipage}{0.245\textwidth}
    \centering \scriptsize{(b)}
\end{minipage} \\
\begin{minipage}{0.162\textwidth}
{\includegraphics[trim={1bp 7bp 50bp 18bp},clip,width=\linewidth]{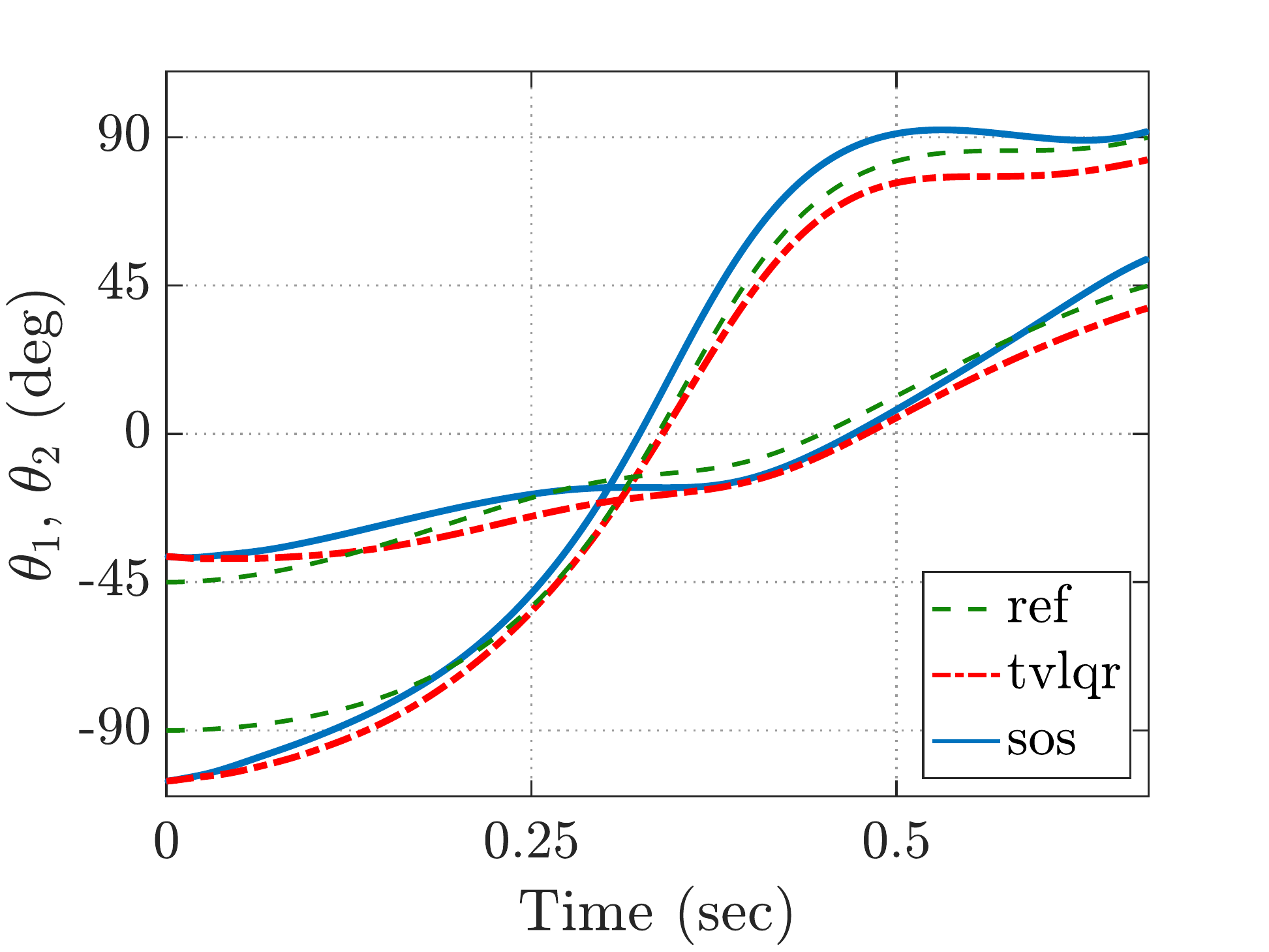}}
\end{minipage}%
\begin{minipage}{0.162\textwidth}
{\includegraphics[trim={0bp 7bp 50bp 18bp},clip,width=\linewidth]{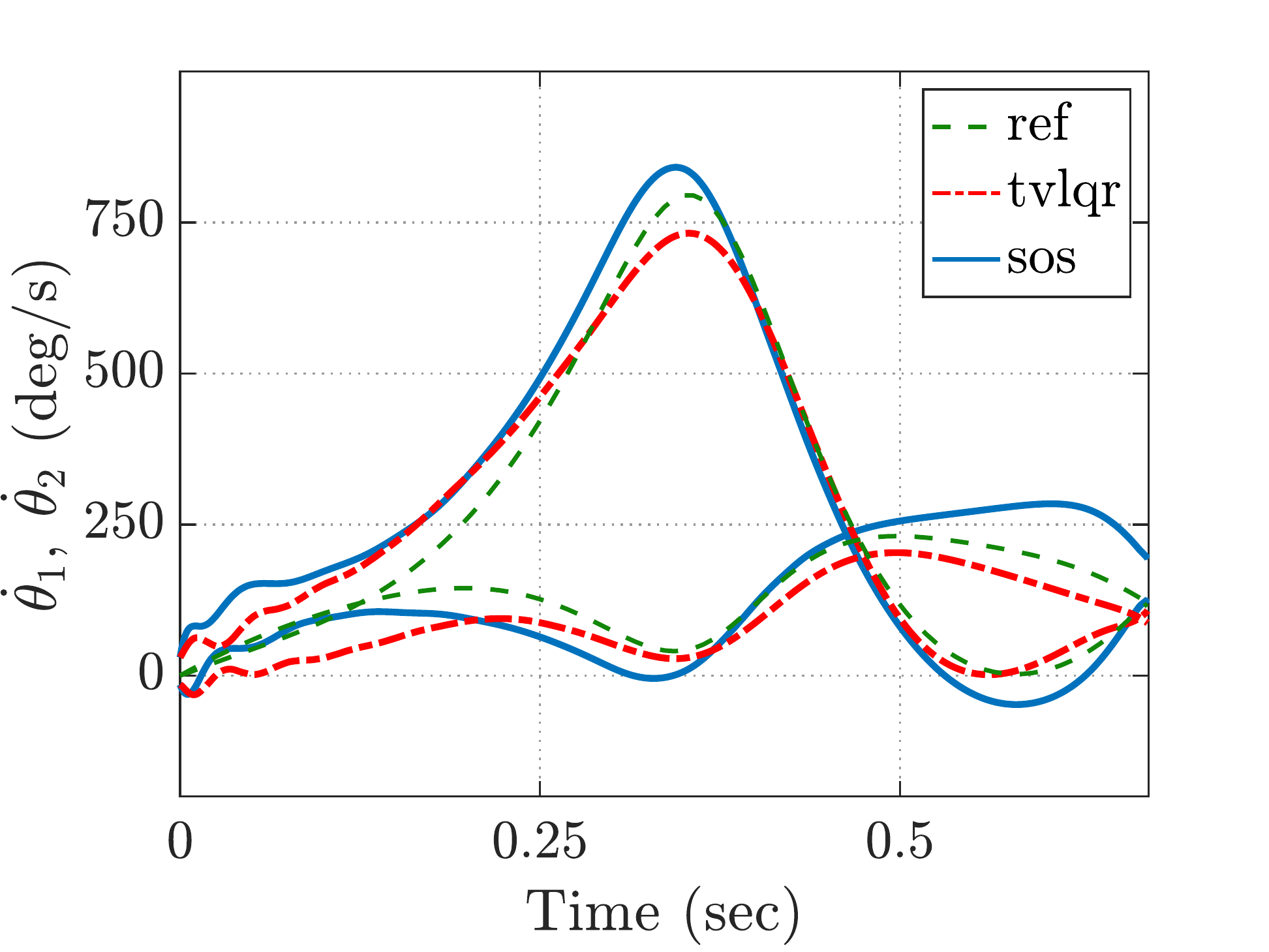}}
\end{minipage}%
\begin{minipage}{0.162\textwidth}
{\includegraphics[trim={4bp 7bp 50bp 18bp} ,clip,width=\linewidth]{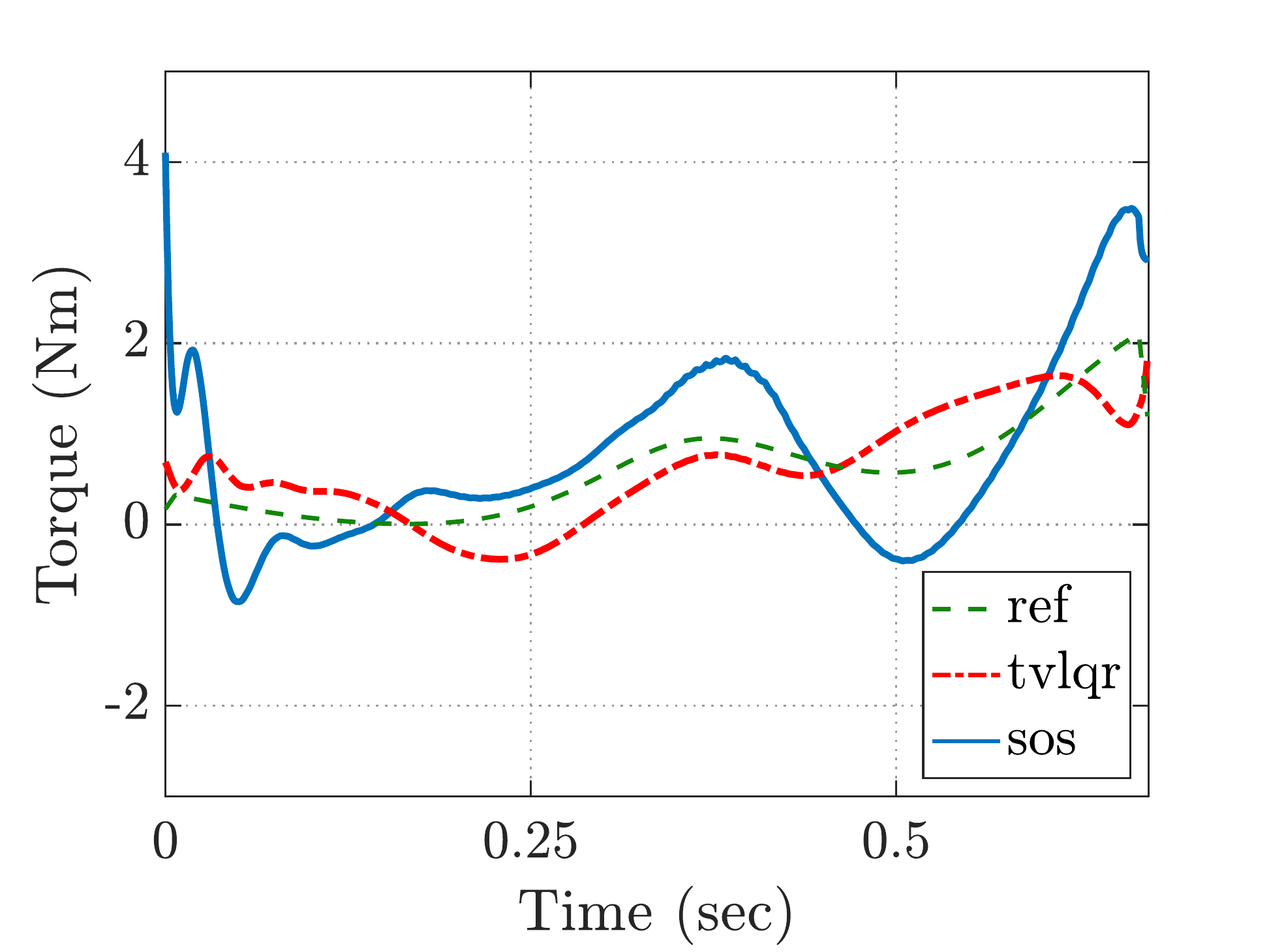}}
\end{minipage} \\
\begin{minipage}{.162\textwidth}
    \centering \scriptsize{(c)}
\end{minipage}%
\begin{minipage}{0.162\textwidth}
    \centering \scriptsize{(d)}
\end{minipage}%
\begin{minipage}{0.162\textwidth}
    \centering \scriptsize{(e)}
\end{minipage}
\vspace{-15pt}
\caption{\label{fig:simulation-motion} Brachiation on flexible cable with $20\%$ stiffness error, starting from off-nominal initial configurations: (a) TVLQR motion trajectory, (b) SOS-based motion trajectory, (c) joint trajectories, (d) joint velocities, (e) torque profiles.}
\end{figure}

Fig. \ref{fig:simulation-motion} shows one of the above experiments, where the robot starts on a cable with stiffness of $20\%$ less than the nominal value and from the off-nominal initial configuration of $[-42^{\circ},\,-100^{\circ},1.85\,\textrm{m},$ $-10\,\textrm{deg/s},\,20\,\textrm{deg/s},\,0.1\,\textrm{m/s}]$.
The results of the TVLQR, and the SOS-based feedback controllers are compared in Fig. \ref{fig:simulation-motion}.
As shown in Fig. \ref{fig:simulation-motion}(c)-(d), the states of the system under SOS-based feedback control successfully track the reference trajectory and approach the desired final configuration, with the final joint angles of $[47.23^{\circ}, \, 91.86^{\circ}]$ and joint velocities of $[192.1, \, 127.2]$ deg/s. The control torque input resulted by SOS-based control is in the range of $[-0.9,\,4.1]$ Nm (Fig. \ref{fig:simulation-motion}(e)), which is within the joint torque limits of $\pm 5$ Nm. However, the TVLQR controller under the same conditions did not succeed in approaching the desired final configuration, ending up at the joint angles of $[41.24^{\circ}, \, 83.25^{\circ}]$ degrees and the joint velocities of $[135.45, \, -110.18]$ deg/s.

\subsection{Continuous Brachiation over Cable via Trajectory Library}
To traverse the entire length of a cable, the robot needs to perform ``continuous'' brachiation, a chained locomotion sequence in which the robot conducts multiple, sequential swings. This scenario presents a real-world challenge since the robot starts from non-zero dynamic states for all except the first swing, and the cable continues to vibrate significantly throughout the motion.

\begin{figure}[t]
\begin{minipage}{0.162\textwidth}
{\includegraphics[trim={50bp 5bp 105bp 30bp},clip,width=\linewidth]{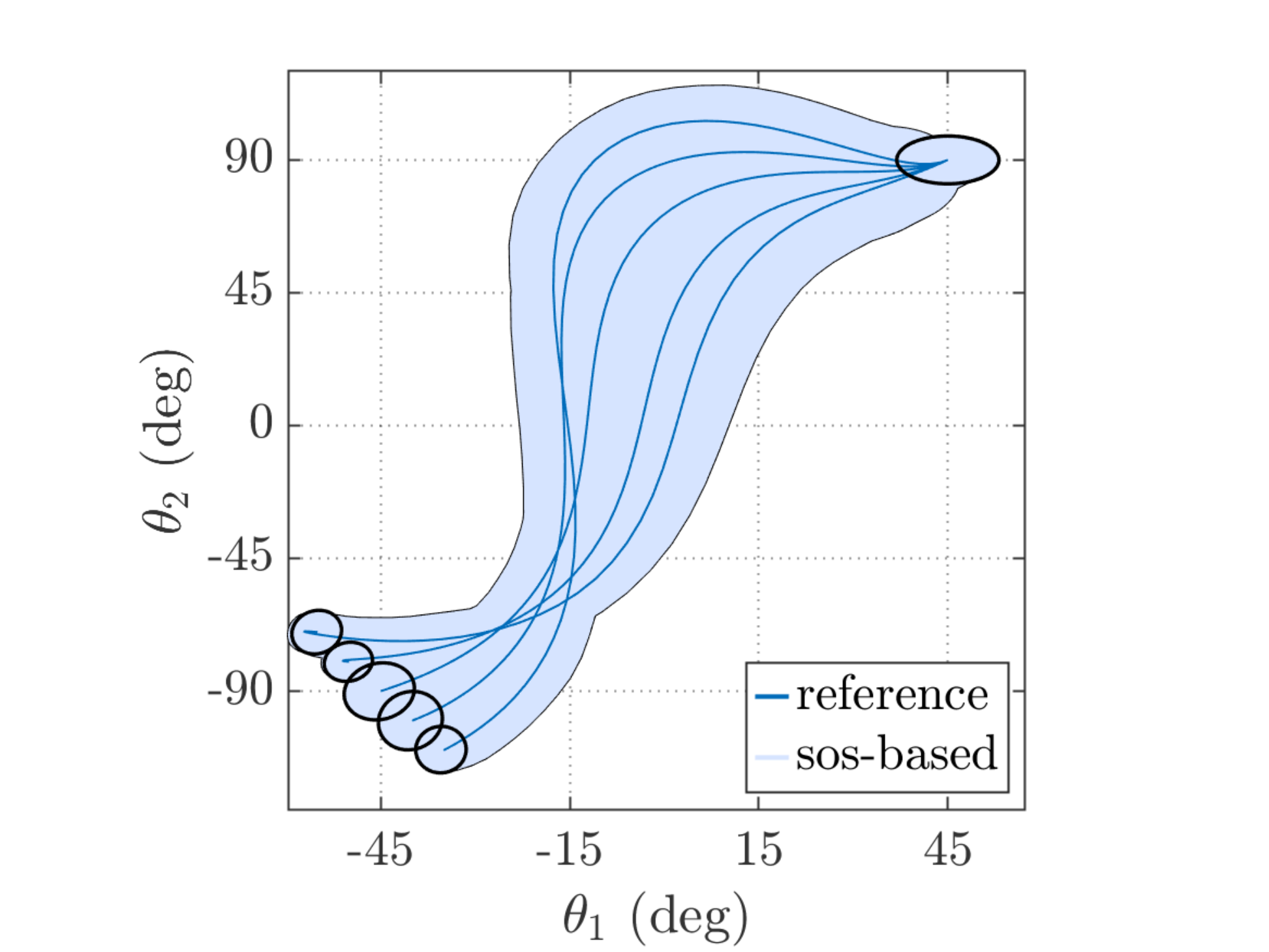}}
\end{minipage}%
\begin{minipage}{0.162\textwidth}
{\includegraphics[trim={50bp 5bp 105bp 30bp},clip,width=\linewidth]{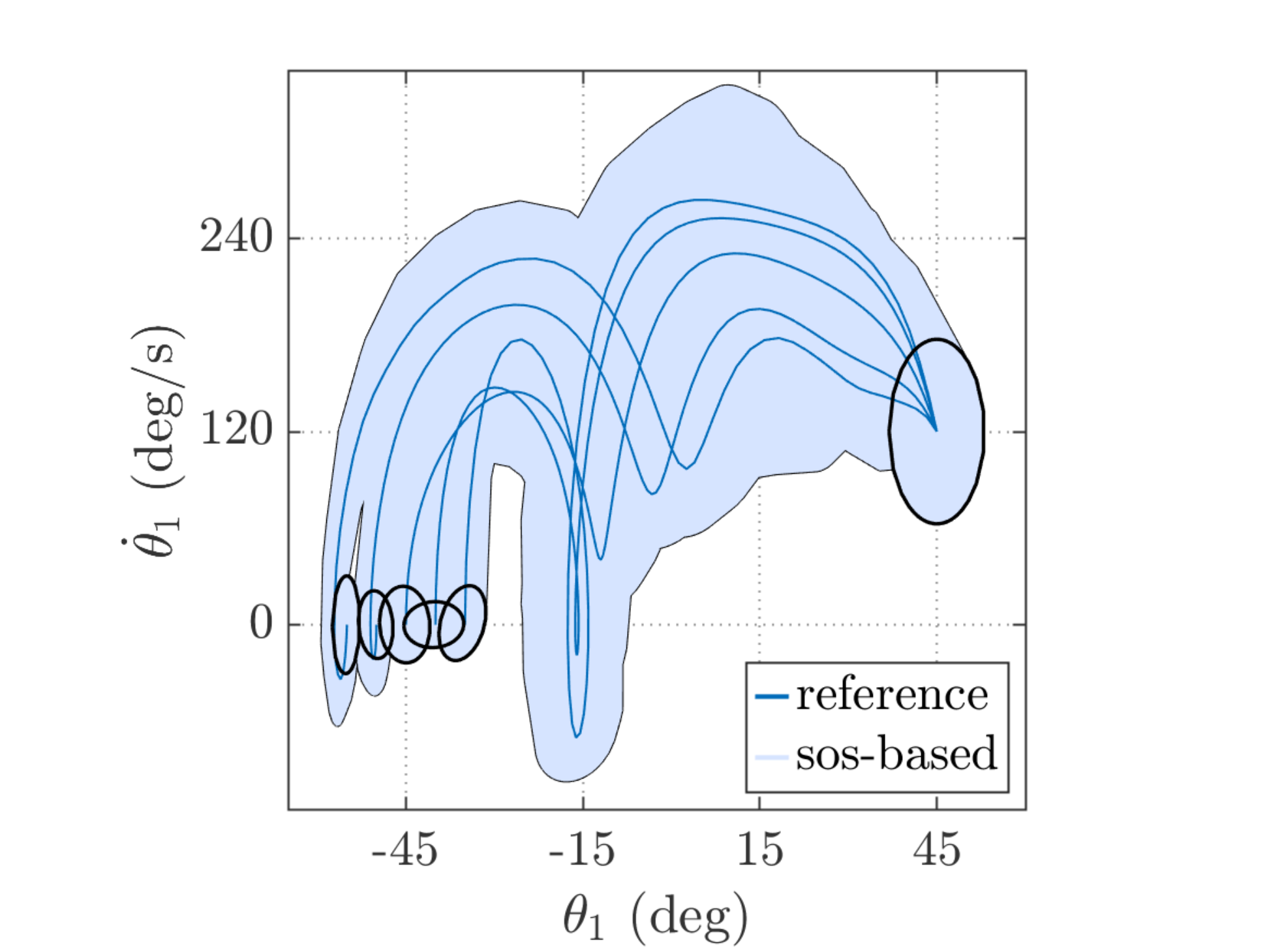}}
\end{minipage}%
\begin{minipage}{0.162\textwidth}
{\includegraphics[trim={40bp 5bp 105bp 30bp}, clip,width=\linewidth]{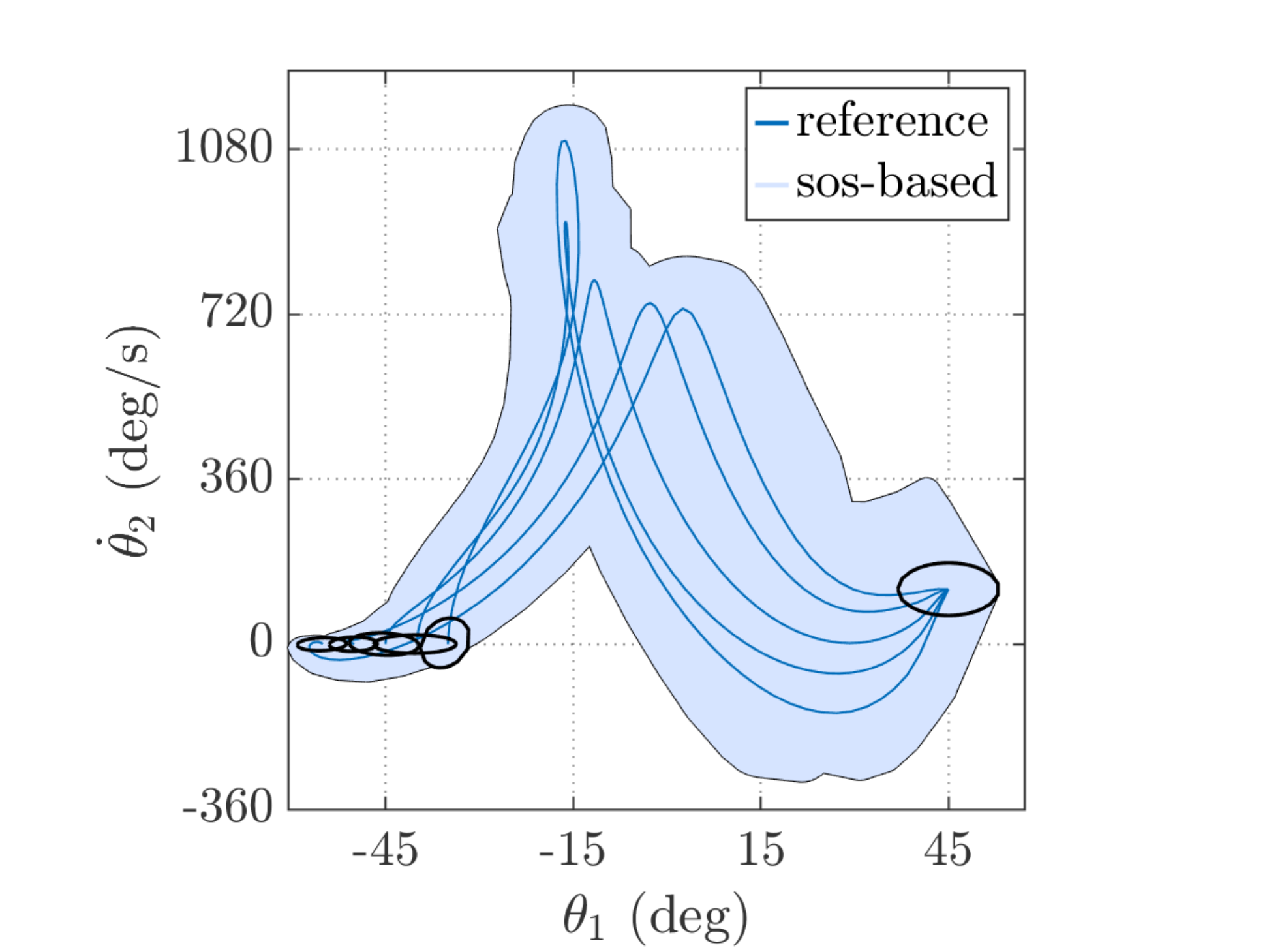}}
\end{minipage} \\
\begin{minipage}{.162\textwidth}
    \centering \scriptsize{(a)}
\end{minipage}%
\begin{minipage}{0.162\textwidth}
    \centering \scriptsize{(b)}
\end{minipage}%
\begin{minipage}{0.162\textwidth}
    \centering \scriptsize{(c)}
\end{minipage} \\
\begin{minipage}{0.2425\textwidth}
{\includegraphics[trim={10bp 50bp 50bp 90bp},clip,width=\linewidth]{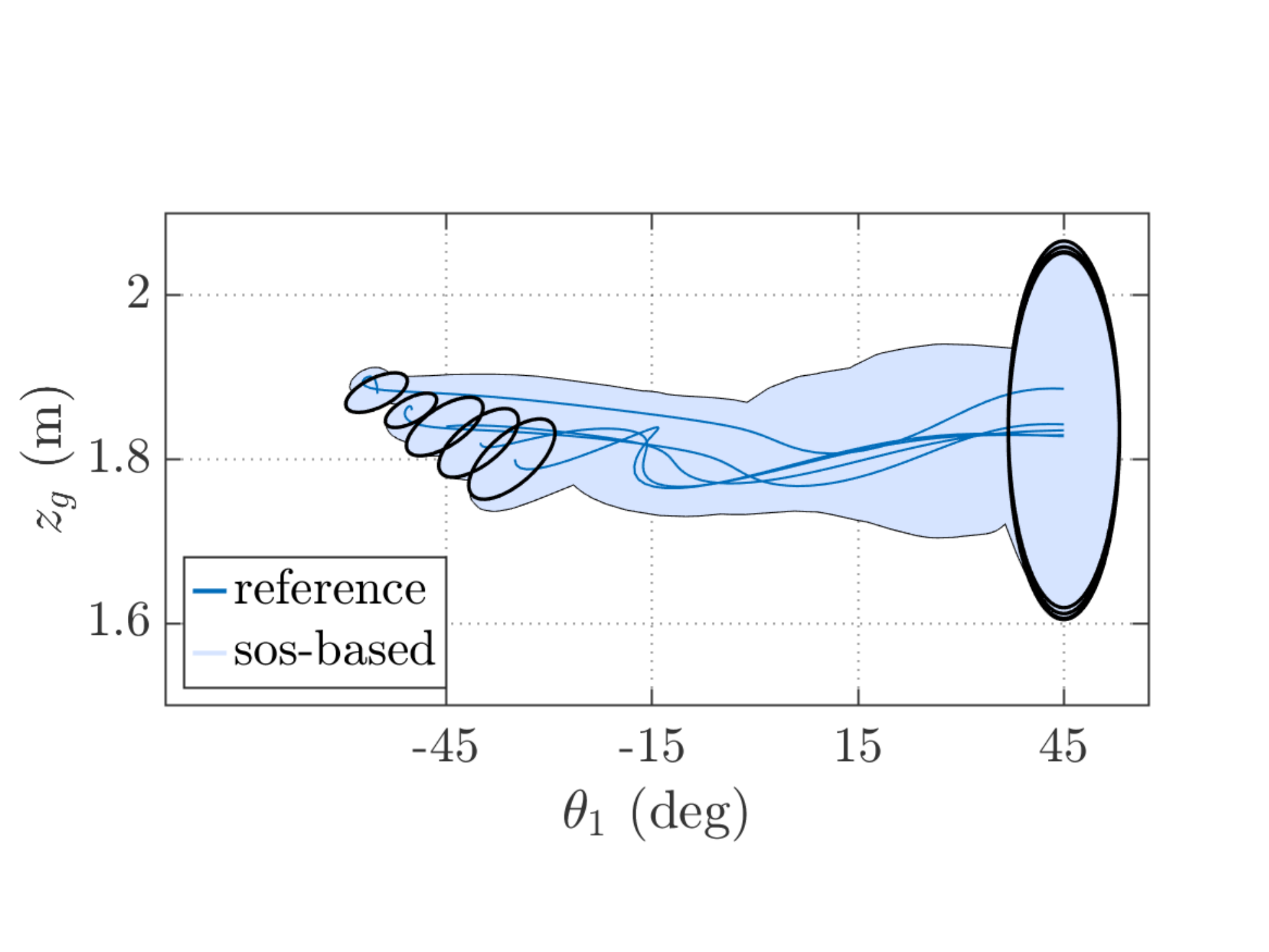}}
\end{minipage}%
\begin{minipage}{0.2425\textwidth}
{\includegraphics[trim={20bp 50bp 50bp 90bp},clip,width=\linewidth]{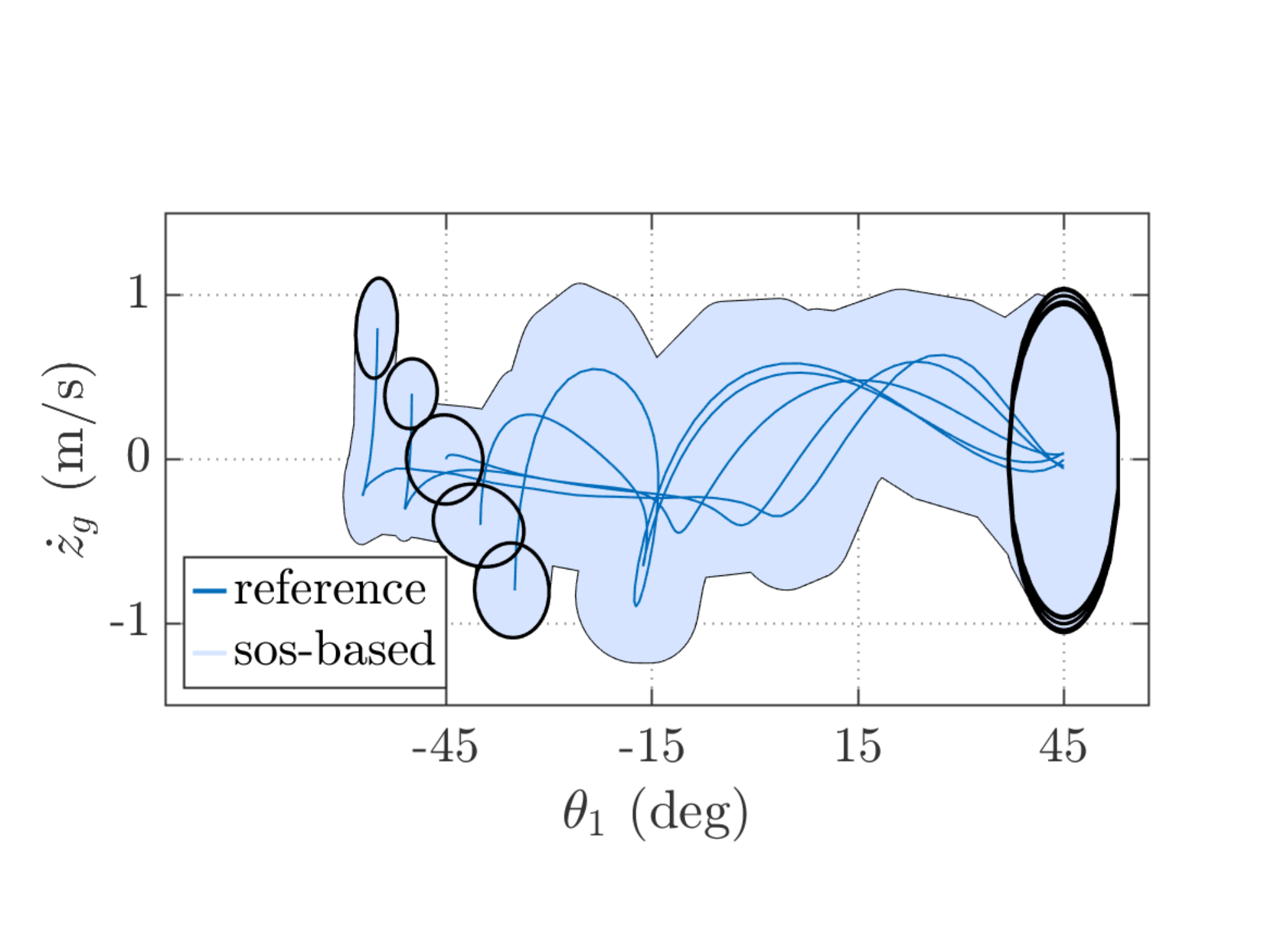}}
\end{minipage} \\
\begin{minipage}{0.2425\textwidth}
    \centering \scriptsize{(d)}
\end{minipage}%
\begin{minipage}{0.2425\textwidth}
    \centering \scriptsize{(e)}
\end{minipage}
\vspace{-15pt}
\caption{The inner-approximation of the backward reachable sets for the trajectory library (displaying 5 out of 10 trajectories). Projection of $\theta_1$ vs. (a) $\theta_2$, (b) $\dot{\theta}_1$, (c) $\dot{\theta}_2$, (d) $z_g$, (e) $\dot{z}_g$.} \label{fig:library} \vspace{-8pt}
\end{figure}

To perform continuous brachiation on the flexible cable, we form a trajectory library containing 10 optimal trajectories. The trajectories start from a range of initial joint angles ($\theta_1,\,\theta_2$) on the cable, and a combination of pivot gripper positions and velocities ($z_g,\,\dot{z}_g$) to account for the cable vibration, but all with zero initial joint velocities ($\dot{\theta}_1,\,\dot{\theta}_2$), as the robot starts each swing with both grippers attached to the cable. A few of these trajectories and their verified invariant regions associated to their SOS-based robust controllers are plotted in Fig. \ref{fig:library}. The final states of all optimal trajectories are set to the desired configuration of $[45^{\circ},90^{\circ},1.9\,\textrm{m},120\,\textrm{(deg/s)},120\,\textrm{(deg/s)},0]$.

\begin{figure}[t]
\renewcommand{\arraystretch}{0.25}
\begin{tabular}{c}
\includegraphics[trim={155bp 7bp 175bp 30bp}, clip,width=0.92\columnwidth]{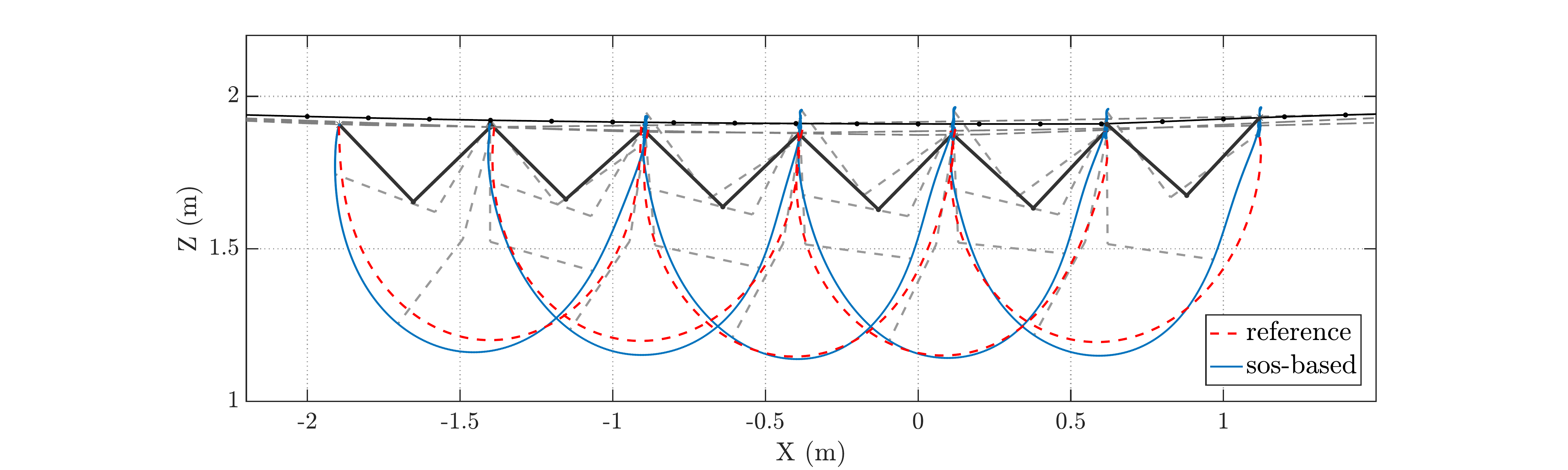}
\tabularnewline
{\scriptsize{}(a)}
\tabularnewline
\includegraphics[trim={100bp 0bp 115bp 20bp}, clip,width=0.96\columnwidth]{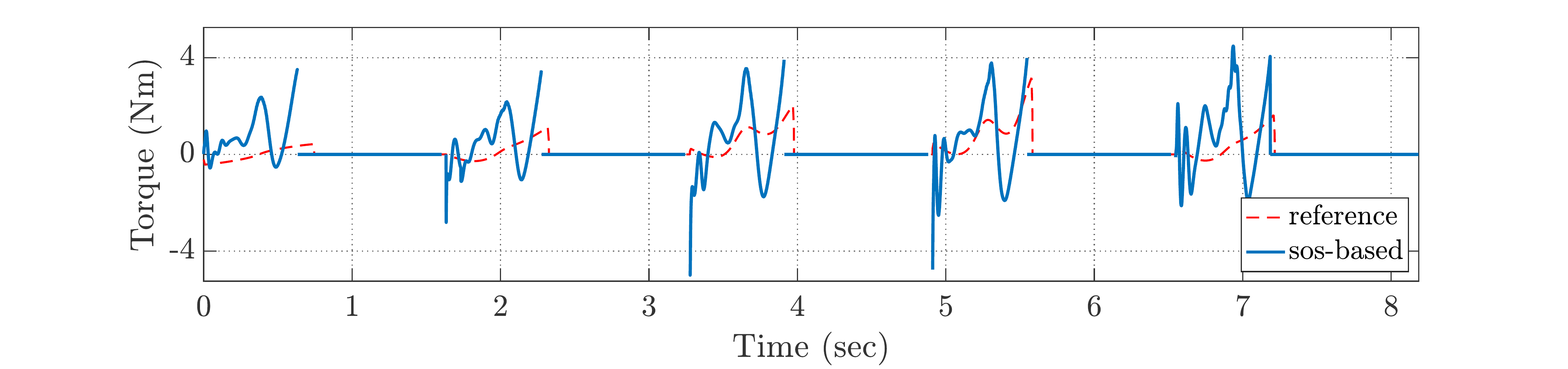}
\tabularnewline
{\scriptsize{}(b)}
\tabularnewline
\end{tabular}
\caption{\label{fig:continuous} Continuous brachiation on flexible cable with $20\%$ stiffness error, using a trajectory library and their SOS-based controllers (a) Feedback motion trajectory, (b) Torque profile.}
\end{figure}

Based on the initial configuration of the robot at the beginning of each swing, the controller framework chooses one of the trajectories and its associated controller to be applied over the next brachiation motion. The resulting motion trajectories and the resulting torque under the SOS-based controllers are shown in Fig. \ref{fig:continuous}. Note that there is only a pause of 1 second enforced between sequential swings. While the reference trajectories are not designed for the exact configuration of each swing, the robust SOS-based feedback controller enables the robot to reliably traverse the length of the cable in 5 successful swings.
As can be seen on Fig. \ref{fig:continuous}(a), for all the swings under the SOS-based controller, the free gripper approaches the cable with the desired final joint angles of $[45^{\circ},90^{\circ}]$, which is considered a successful motion for this application.

\vspace{-5pt}
\subsection{Validation by Hardware Experiments}
\begin{figure}[t]
\centering
\frame{\includegraphics[trim={0bp 0bp 0bp 0bp},clip, width=0.7\columnwidth]{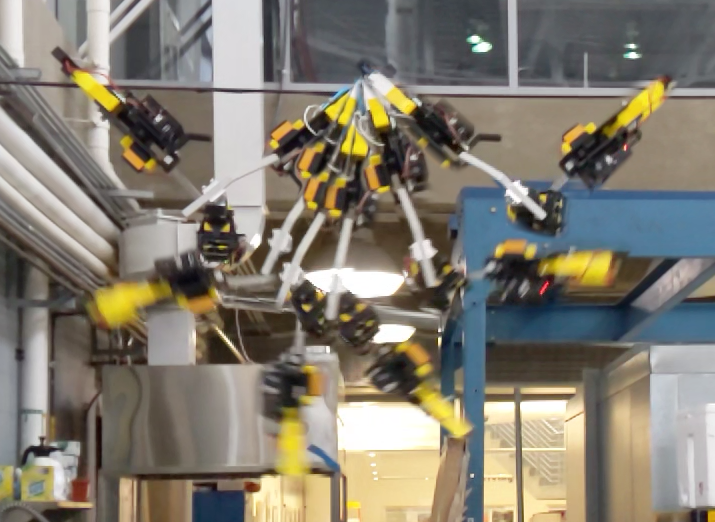}}
\vspace{-2pt}
\caption{Time-lapse of the robot brachiation traveling from left to right.}
\label{fig:experiment-timelapse}\vspace{-8pt}
\end{figure}

\vspace{-2pt}
Experimental validation of the developed robust feedback controller has been conducted on the wire-borne brachiating robot shown in Fig. \ref{fig:robot-hardware}.
Fig. \ref{fig:experiment-timelapse} shows a time-lapse of six frames in a successful brachiation maneuver in an experimental setting. The overall robot path is similar to the trajectories identified in simulation (Fig. \ref{fig:continuous}).\looseness=-1

\begin{figure}[t]
\vspace{-2pt}
\begin{minipage}{0.25\textwidth}
    \centering {\includegraphics[trim={60bp 2bp 102bp 28bp}, clip,width=\linewidth]{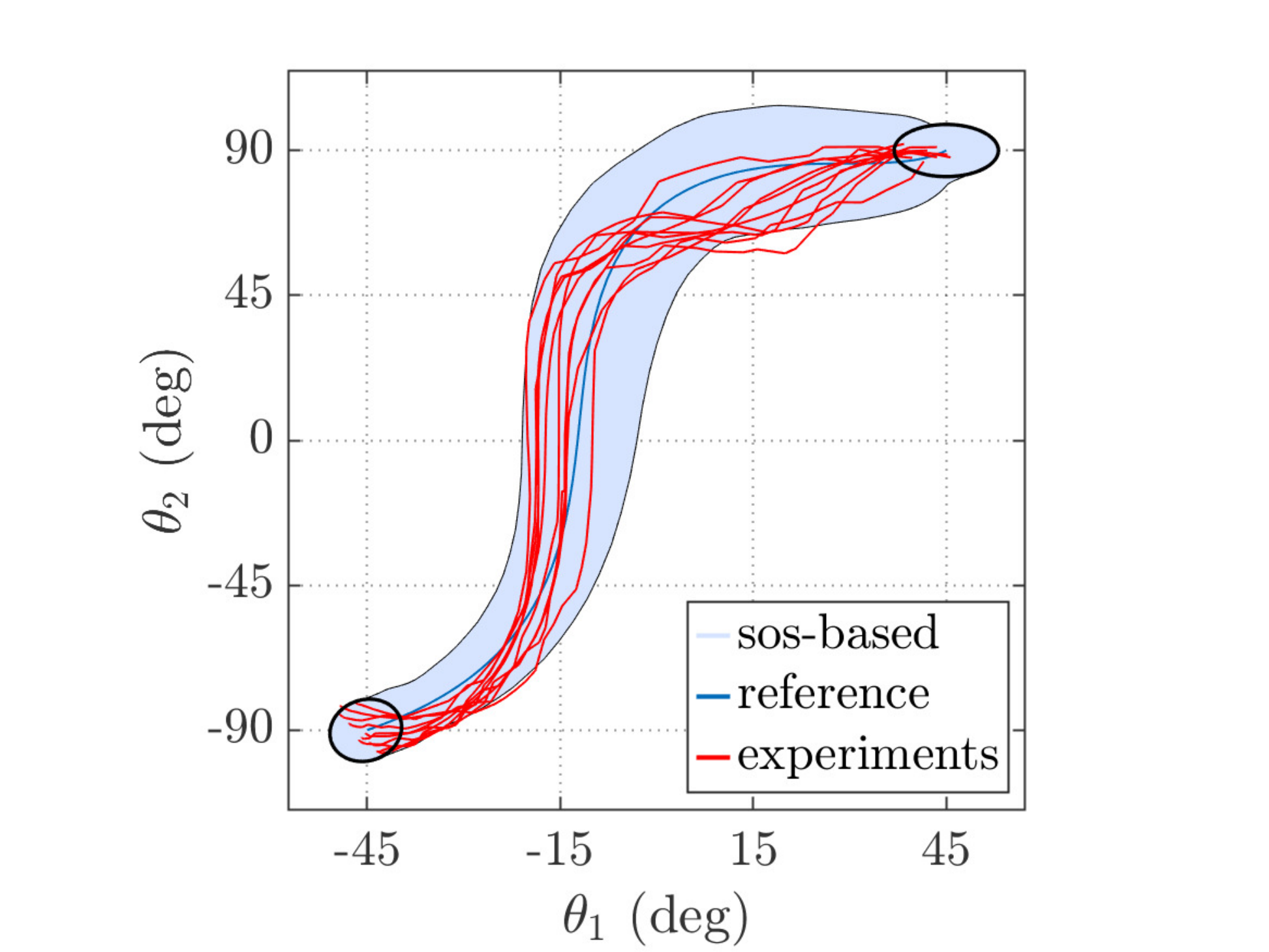}}
\end{minipage}%
\begin{minipage}{0.22\textwidth}
\centering
\includegraphics[trim={0bp 50bp 45bp 90bp}, clip,width=\linewidth]{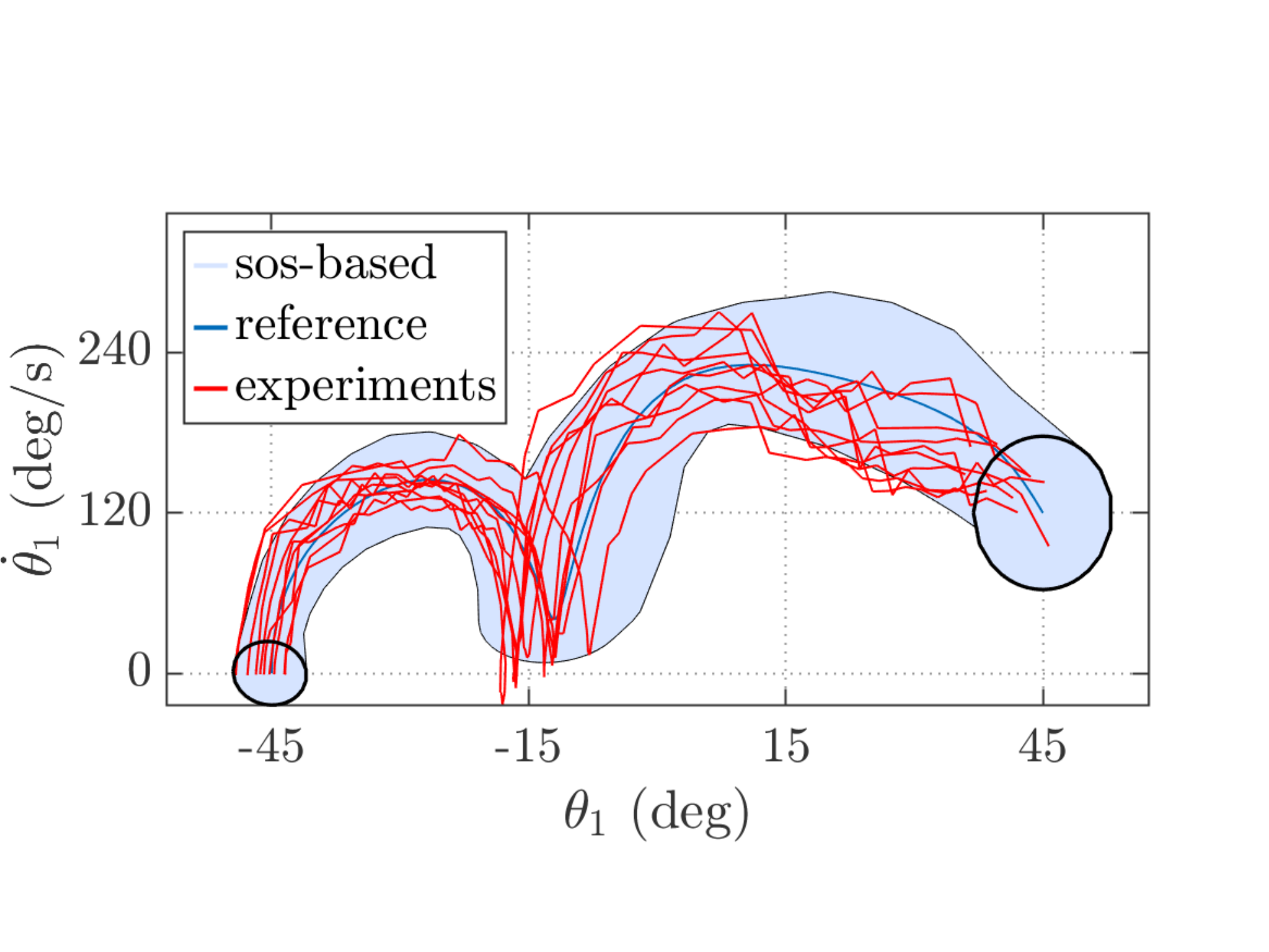} \\
\includegraphics[trim={0bp 50bp 45bp 90bp}, clip,width=\linewidth]{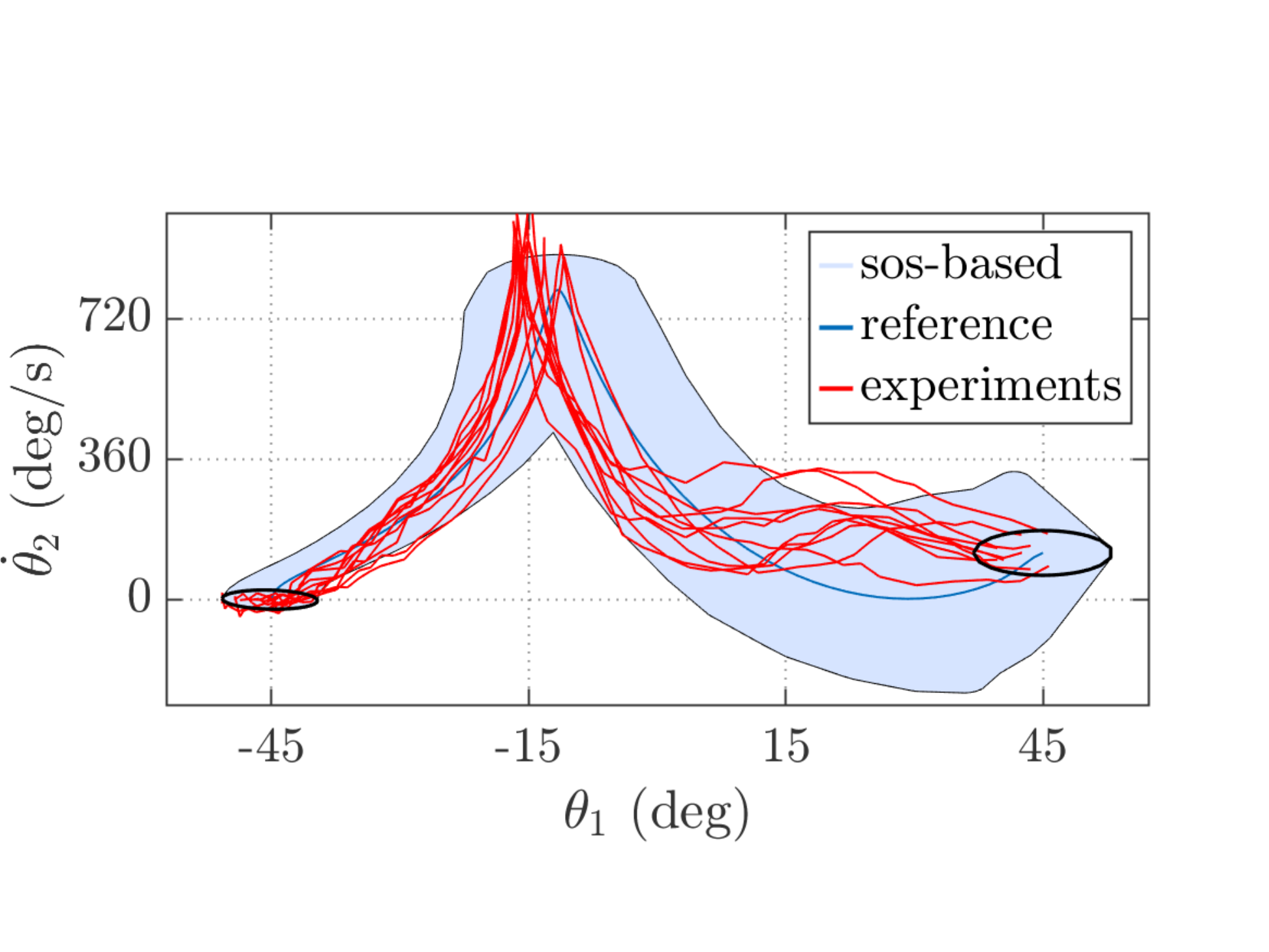}
\end{minipage} \\
\begin{minipage}{0.25\textwidth}
\centering \scriptsize{(a)}
\end{minipage}%
\begin{minipage}{0.22\textwidth}
\centering \scriptsize{(b) , (c)}
\end{minipage}
\vspace{-6pt}
\caption{Experimental motion trajectories with the SOS-based controller plotted on top of the approximated backward reachable set. Projection of $\theta_1$ vs. (a) $\theta_2$, (b) $\dot{\theta}_1$, (c) $\dot{\theta}_2$.} \label{fig:invset-experiment}
\end{figure}

Data was collected from 10 closed-loop control experiments of the hardware brachiating robot attached to an 8-meter flexible cable. Using a single nominal trajectory, and starting from various initial conditions on the cable within the verified set of initial conditions, the time-varying SOS-based controller was applied for the time horizon of the nominal trajectory. Fig. \ref{fig:invset-experiment} plots the resulting motion trajectories on top of the projections of the robust verified backward reachable set. Despite initial conditions and cable stiffness different from the nominal trajectory and values, the controller performs successfully and reach the desired final configuration. While some of the experimental trajectories slightly violate the verified regions, this could be due to the conservative and inner-approximation formulation used to derive the invariant sets, as well as the inherent model mismatch between the dynamic model and the actual hardware. Note that the gripper states ($z_g$ and $\dot{z}_g$) are unmeasurable states of the system.\looseness=-1


\begin{figure}[b]
\begin{minipage}{0.49\textwidth}
\centering {\begin{tikzpicture}\node[inner sep=0pt] at (0,0)
{\includegraphics[trim={3bp 30bp 35bp 20bp}, clip,width=0.5\linewidth]{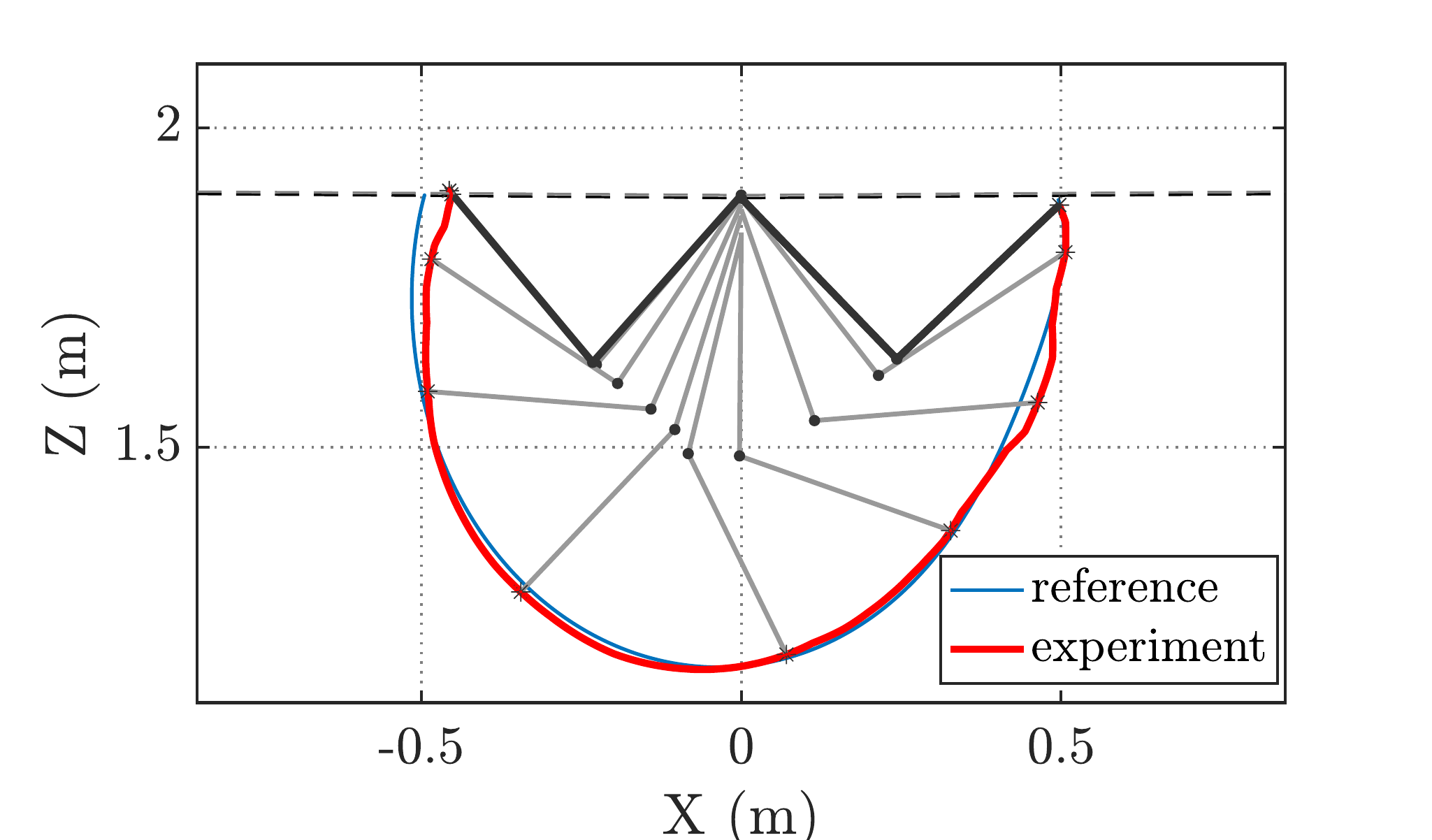}};
\draw[<-,>=stealth',semithick,red,dashed] (-0.9,-0.2) to [out=130,in=230] (-0.9,0.65);
\end{tikzpicture}}
\end{minipage} \\
\begin{minipage}{0.49\textwidth}
    \centering \scriptsize{(a)}
\end{minipage} \\
\begin{minipage}{0.162\textwidth}
\centering {\includegraphics[trim={1bp 7bp 50bp 18bp},clip,width=0.95\linewidth]{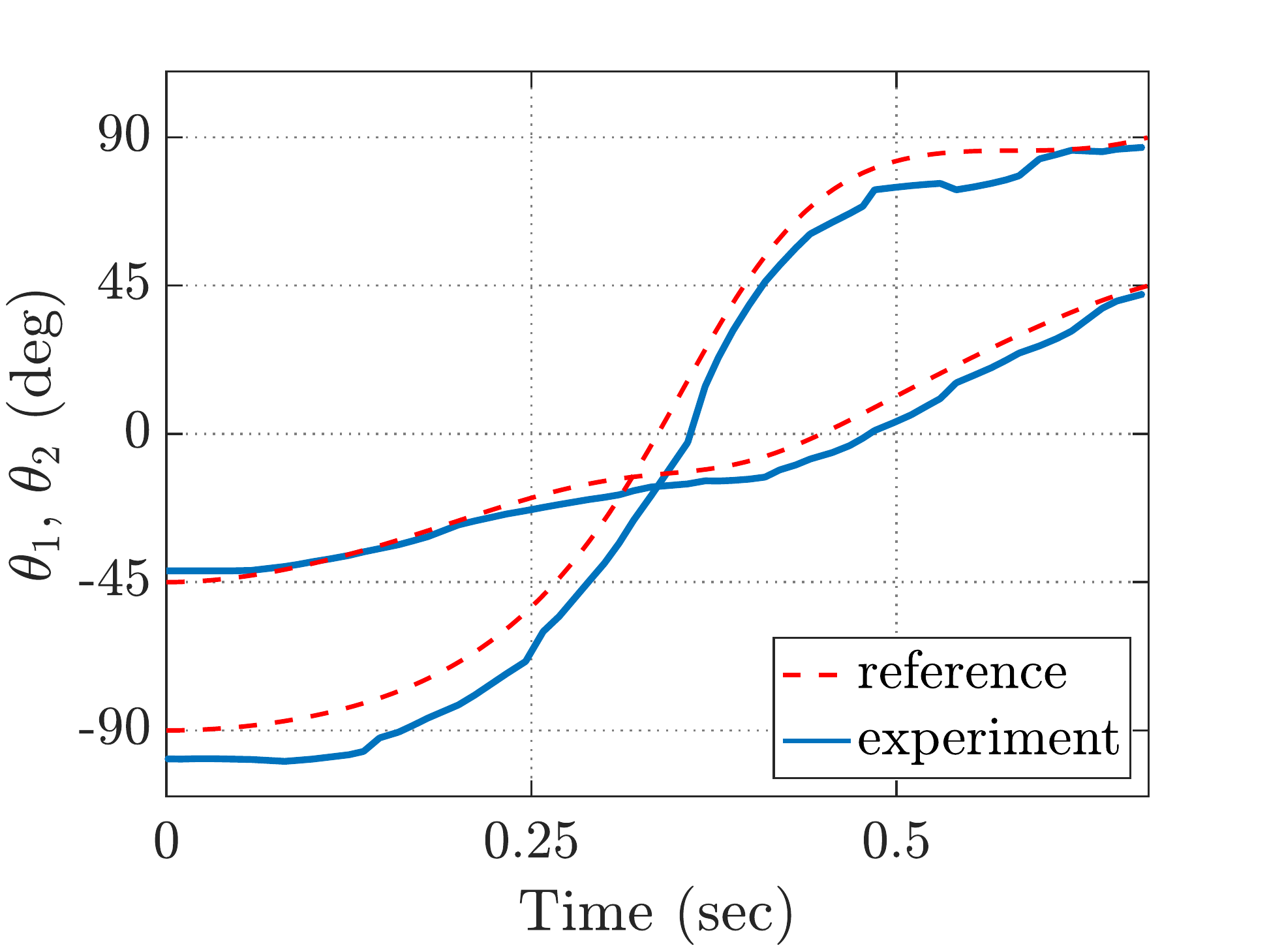}}
\end{minipage}%
\begin{minipage}{0.162\textwidth}
 {\includegraphics[trim={0bp 7bp 50bp 18bp},clip,width=0.95\linewidth]{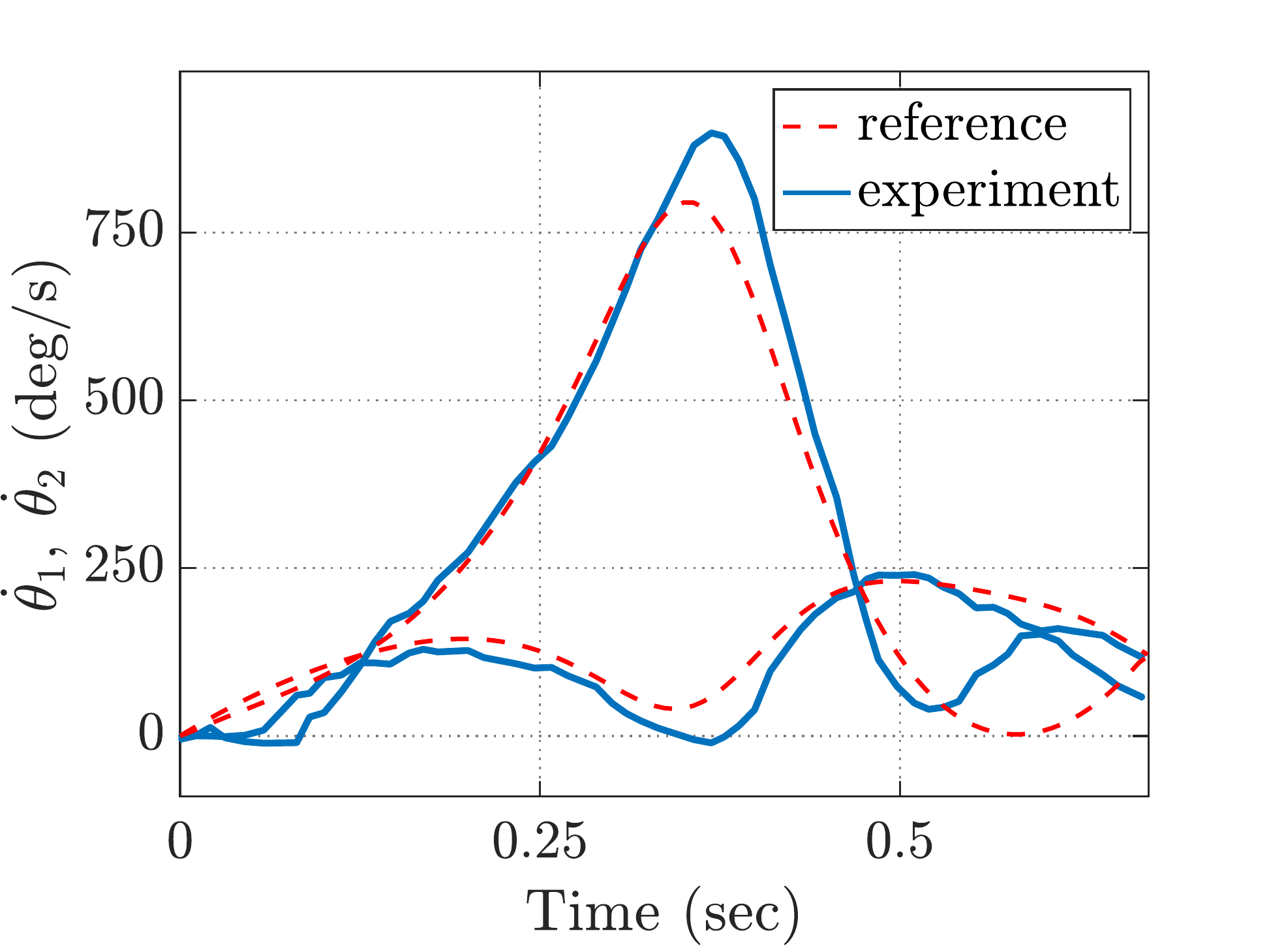}}
\end{minipage}%
\begin{minipage}{0.162\textwidth}
\centering {\includegraphics[trim={20bp 7bp 50bp 18bp} ,clip,width=0.95\linewidth]{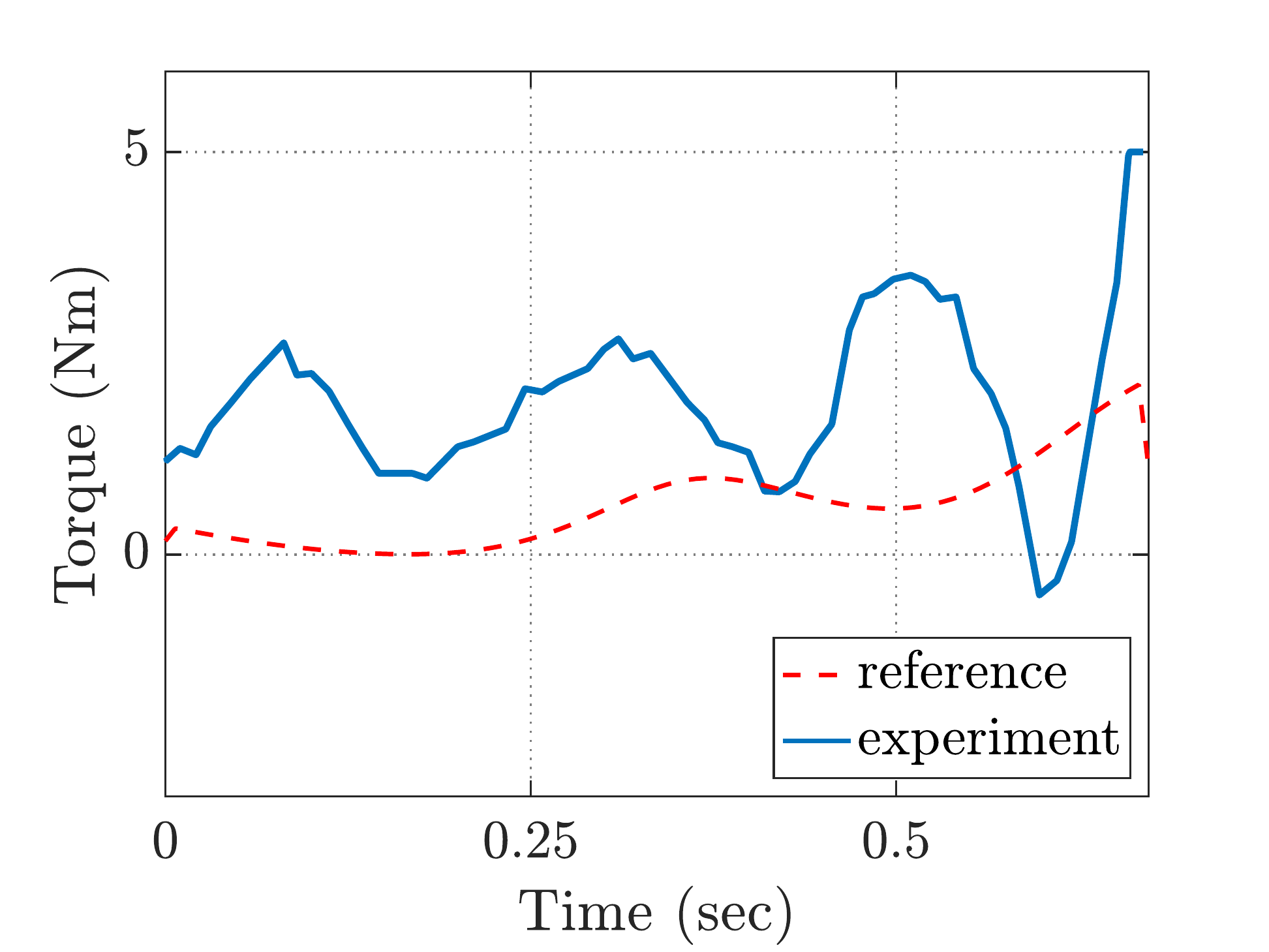}}
\end{minipage} \\
\begin{minipage}{0.162\textwidth}
    \centering \scriptsize{(b)}
\end{minipage}%
\begin{minipage}{0.162\textwidth}
    \centering \scriptsize{(c)}
\end{minipage}%
\begin{minipage}{0.162\textwidth}
    \centering \scriptsize{(d)}
\end{minipage}
\vspace{-18pt}
\caption{An example experimental result for brachiation on flexible cable, starting from off-nominal initial configurations: (a) SOS-based motion trajectory, (b) joint trajectories, (c) joint velocities, (d) torque profile.} \label{fig:experiment-motion}
\end{figure}

A visualization of one of the experimental swings of the physical hardware is shown in Fig. \ref{fig:experiment-motion}. The robot starts on the cable from the off-nominal initial configuration of $[-42.5^{\circ},\,-99^{\circ},1.85\,\textrm{m},\,0,\,0,\,0]$, and successfully tracks the reference trajectory and approaches the desired final configuration, with final joint angles of $[43^{\circ}, \, 88^{\circ}]$ and joint velocities of $[116, \, 57]$ deg/s. 
The hardware controller has only minor deviations from the reference trajectory while achieving a final configuration with low positional and velocity errors. Additionally, the controller does not saturate the actuator, with control torque input in the range of $[-0.5,\,5]$ Nm.\looseness=-1

The videos of the simulation results and the hardware experiments presented in this section can be found in the accompanying video of the paper, available at {\small \url{https://vimeo.com/sfarzan/iros20}}.

\section{CONCLUSIONS}
We presented and experimentally tested a robust time-varying feedback controller with formal guarantees for a two-link underactuated brachiating robot traversing flexible cables. A simplified dynamic model for flexible cables is proposed, in which the vibration dynamics of the cable is 
closely approximated by parallel spring-dampers attached to different heights,
providing the ability to include parametric model uncertainties caused by the flexible cable.
Semidefinite optimization and sum-of-squares programming are used to compute an inner-approximation to the backward reachable set around an optimal trajectory for a given set of desired final configurations. A robust SOS-based feedback control is synthesized to maximize the size of the reachable set, while considering the bounded parametric model uncertainties, actuator saturations and unmeasurable states in the system. A library of optimal trajectories and their associated SOS-based controllers is formed to enable the robot to traverse the entire length of the cable in a continuous fashion.\looseness=-1

The resulting verified regions, as well as the reported simulation and hardware experiments, demonstrate that compared to a TVLQR controller, the SOS-based controller results in larger robust backward reachable sets, enabling the robot to employ fewer reference trajectories and associated controllers to reliably traverse a flexible cable under model uncertainty.

\vspace{-4pt}
\bibliographystyle{IEEEtran}
\bibliography{sfarzan_bibtex_0919}

\end{document}